\documentclass[amsfonts, amssymb, amsmath, reprint, showkeys, nofootinbib, twoside, floatfix]{revtex4-2}
\usepackage{graphicx}
\usepackage{xcolor}
\usepackage{placeins}
\usepackage[english]{babel}
\usepackage{svg}
\usepackage[utf8]{inputenc}
\usepackage{float}
\usepackage{soul}
\usepackage{ulem}
\usepackage{amsthm}
\usepackage{mathtools}
\usepackage{physics}
\usepackage{xcolor}
\usepackage{graphicx}
\usepackage[left=23mm,right=13mm,top=35mm,columnsep=15pt]{geometry} 
\usepackage{adjustbox}
\usepackage{placeins}
\usepackage[T1]{fontenc}
\usepackage{lipsum}
\usepackage{csquotes}

\usepackage{bbm}

\usepackage[colorinlistoftodos, color=green!40, prependcaption]{todonotes}
\usepackage[pdftex, pdftitle={Article}, pdfauthor={Author}]{hyperref} %
\usepackage{hyperref}

\newboolean{showcomments}
\setboolean{showcomments}{true}
\newcommand{\HL}[1]{\ifthenelse{\boolean{showcomments}}{{\color{red}{HL: #1}}}{}}
\newcommand{\SM}[1]{\ifthenelse{\boolean{showcomments}}{{\color{blue}{SM: #1}}}{}}
\newcommand{\BA}[1]{\ifthenelse{\boolean{showcomments}}{{\color{magenta}{BA: #1}}}{}}
\newcommand{\BAb}[1]{\ifthenelse{\boolean{showcomments}}{{\color{black}{BA: #1}}}{}}
\newcommand{\sm}[1]{\ifthenelse{\boolean{showcomments}}{{\color{green}{SM: #1}}}{}}
\begin{document}
\title{Contextual Quantum Neural Networks for Stock Price Prediction}

\author{Sharan Mourya}
    \affiliation{Department of Electrical and Computer Engineering,
University of Illinois at Urbana-Champaign}
    \affiliation{Fujitsu Research of America}
\author{Hannes Leipold}
    \affiliation{Fujitsu Research of America}
\author{Bibhas Adhikari}
    \affiliation{Fujitsu Research of America}
\date{\today}

\begin{abstract}
In this paper, we apply quantum machine learning (QML) to predict the distribution of stock prices of multiple assets using a contextual quantum neural network. Our approach captures recent trends to predict future stock price distributions, moving beyond traditional models that focus on entire historical data. Utilizing the principles of quantum superposition, we introduce a new training technique called the quantum batch gradient update (QBGU), which accelerates the standard stochastic gradient descent (SGD) in quantum applications and improves convergence. Consequently, we propose a quantum multi-task learning (QMTL) architecture, specifically, the share-and-specify ansatz, that integrates task-specific operators controlled by quantum labels, enabling the simultaneous and efficient training of multiple assets on the same quantum circuit as well as enabling efficient portfolio representation with logarithmic overhead in the number of qubits. Through extensive experimentation on S\&P 500 data for Apple, Google, Microsoft, and Amazon stocks, we demonstrate that our approach outperforms quantum single-task learning (QSTL) models by effectively capturing inter-asset correlations. Our findings highlight the transformative potential of QML in financial applications, paving the way for more advanced, resource-efficient quantum algorithms in stock price prediction and other complex financial modeling tasks.
\end{abstract}


\maketitle
\raggedbottom 

\section{Introduction}

Quantum computing is a computational paradigm that transcends the limitations of classical computing by harnessing the principles of quantum superposition and entanglement. These unique features enable quantum computers to tackle complex problems faster than classical systems can\cite{shor,hhl,aharonov2006polynomial,rebentrost2014quantum}. Owing to the potential exponential scaling of computational power with the number of qubits, quantum computing is expected to revolutionize diverse sectors, including medicine, engineering, energy, and finance~\cite{apps,supremacy}. Despite its immense potential, building a fully functional quantum computer is a monumental challenge that could take years, if not decades, to achieve a clear computational advantage over classical computers~\cite{limits,nisq}. However, the near-term applications of quantum computing are particularly promising in fields like finance, where its ability to process vast amounts of complex data with fewer resources is transformative. Even with today's noisy intermediate scale quantum (NISQ) devices - limited by a small number of qubits and short coherence times~\cite{nisq} - quantum methods can provide approximate solutions to certain financial problems, making them highly relevant in the immediate future~\cite{review1}.

Machine learning has become essential for financial tasks like, asset management~\cite{review1}, risk analysis~\cite{credit}, crash detection~\cite{crash}, and portfolio optimization~\cite{fin1}. The ability of machine learning algorithms to analyze massive datasets, recognize patterns, and make fast predictions provides a significant competitive edge. Quantum machine learning (QML)~\cite{qml} emerges at the intersection of these fields, combining quantum computing's ability to process and represent complex states efficiently with the powerful predictive tools of machine learning~\cite{classical}. With the exponential growth of financial data, current machine learning systems are quickly reaching the boundaries of classical computational models. In this context, quantum algorithms present a promising alternative by offering faster or higher quality solutions for specific classes of problems. Additionally, breakthroughs in quantum learning theory suggest that, under certain conditions, there is a provable distinction between classical and quantum learnability~\cite{expressibility}. This implies that problems deemed challenging for classical systems could see significant improvements through the adoption of QML approaches. 

Quantum machine learning can be broadly categorized into two main components: parametric quantum circuit (PQC) optimization and classical-to-quantum information encoding~\cite{review1}. These categories represent two core components of QML algorithms and workflows, each addressing a different aspect of how classical data interacts with quantum systems and how quantum models are trained. PQCs are quantum circuits that contain tunable parameters, interpretable as weights of a quantum neural network. These parameters are adjusted iteratively to minimize or maximize a cost function, similar to how classical machine learning algorithms optimize parameters during training. PQCs can further be classified into two categories: variational quantum circuits (VQC) and quantum neural networks (QNN). VQCs involve a hybrid architecture where a classical computer works alongside a quantum computer in a loop, while QNNs consist of circuit architectures tailored to specific problems. 

Recently, there have been a surge of research interest in these areas. For instance, hybrid architectures using parameterized quantum circuits in combination with classical optimization loops have been successfully deployed for classification tasks~\cite{pqc}, while support vector machines (SVM) have been utilized for data classification~\cite{svm}. In another study, quantum state space was used as the feature space to improve the learnability of QML and achieve quantum advantage in classification tasks~\cite{z}. Quantum versions of machine learning models, such as Boltzmann machines~\cite{qbm}, recurrent neural networks~\cite{qrnn}, generative adversarial neural networks~\cite{qgal}, reinforcement learning~\cite{qrl}, and reservoir computing~\cite{qrc}, have been extensively studied. 

In finance, quantum machine learning has been applied to various tasks, including options pricing~\cite{options}, time-series forecasting~\cite{time-series}, and stock price prediction~\cite{closing}. A notable example is a hybrid architecture developed for financial predictions~\cite{hybrid}, while quantum Wasserstein generative adversarial neural networks were employed for time-series predictions on the S\&P 500~\cite{qwgan}. Unsupervised quantum machine learning has also been explored for clustering and fraud detection~\cite{review2}. On the other hand, significant progress has also been made in loading classical information onto quantum states. For example, quantum adversarial neural networks have been utilized to load random distributions onto quantum circuits using feature maps~\cite{qgan}, and quantum Wasserstein GANs have achieved similar tasks with a gradient penalty, improving performance over previous approaches~\cite{qwgan}. 

While many financial computational problems like portfolio management and risk analysis require training across multiple assets, only few studies have addressed this need. For instance, joint learning of two distributions was achieved in~\cite{joint}, though its direct application to financial problems like stock price prediction remains limited due to computational expense and the reliance on feature maps, which will be further explained in this paper. A related study~\cite{qmtl} applied quantum reservoir computing and multi-task learning~\cite{mtl}, which is hindered by the complexity of quantum reservoir systems, lacking generalization. 

In this paper, we aim to address the challenge of training a parameterized quantum circuit over multiple assets on a single quantum device by utilizing minimal resources and optimizing various components of the training process. The contributions of this work are as follows:

\begin{enumerate}
    \item We adopted fidelity loss over quantum representations of the entire data distribution and employed a training technique, quantum batch gradient update (QBGU), that loads the amplitude encoded context distribution and empirical context-continuation distribution to accelerate convergence and improve convergence quality compared to training by stochastic gradient descent (SGD) through expensive reconstruction of classical distributions.
    \item We analyzed context-based stock price prediction for various companies using several training paradigms, including parameter shift and (simulation supported) backpropagation, across different loss functions. 
    \item We developed a new quantum multi-task learning (QMTL) architecture - \textit{share-and-specify} ansatz - for predicting the stock price distribution over a portfolio of assets. By achieving logarithmic overhead in the number of assets in the portfolio and controllable scaling in the size of the context, we can load highly nontrivial non-static distributions on quantum devices, enabling the usage of potential quantum advantageous algorithms such as approaches utilizing quantum amplitude estimation~\cite{Brassard} like quantum risk analysis~\cite{woerner2019quantum} or other quantum algorithms for speed-up like quantum linear system solving~\cite{hhl} at inference time.
\end{enumerate}

This paper is organized as follows: Section II gives a brief background of the required terminology used throughout this paper. Section III gives an overview of loading classical data onto quantum registers. Section IV and V introduces quantum single-task and multi-task learning respectively followed by numerical simulations in Section VI.

\section{Background}

\begin{figure*}[t]
\centering
\includesvg[width=1.0\linewidth]{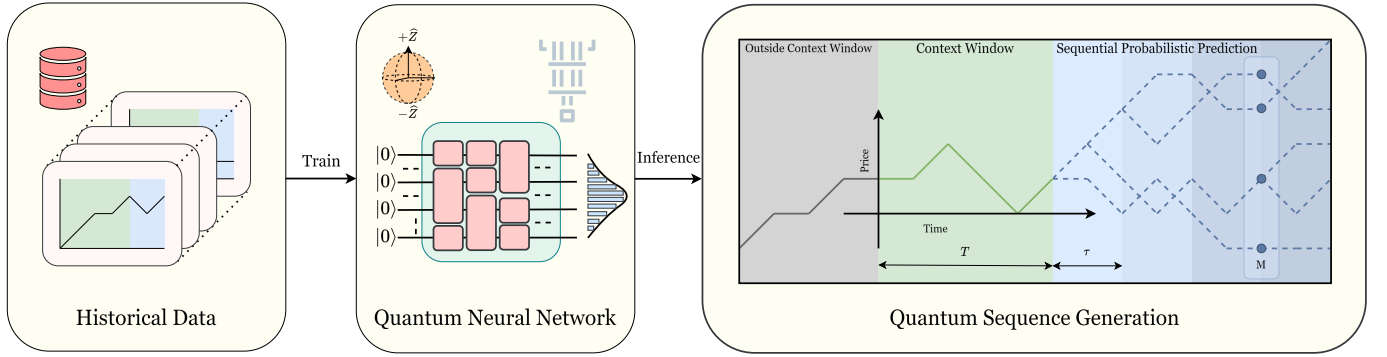}
\caption{\textbf{Quantum Neural Networks for Contextual Sequence Generation}. Given historical data of context and continuations, a Quantum Neural Network is trained to produce quantum distributions over future prices, enabling utilization of quantum advantageous algorithms, such as quantum risk analysis~\cite{woerner2019quantum}, downstream for tasks (labeled $M$ in the rightmost diagram at a particular future) over all sequences in superposition.}
\label{fig:qnnflow}
\end{figure*}

In this section, we introduce all the background information needed to understand this paper including the time-series prediction model and contextual Quantum Neural Networks (QNN) for modeling asset futures.

\subsection{Time-Series Prediction}

A financial asset price prediction model is a time-series forecasting model designed to predict future asset prices by leveraging historical numeric price data and additional contextual information \cite{context}. In this section, we outline the mathematical framework and notations that will be used subsequently to develop a quantum machine learning (QML) model for multi-asset price prediction. For any given financial asset, the associated \textit{contextual string} represents a sequence of numeric values corresponding to the asset over a specific time frame. Typically, this sequence is formed by considering a series of consecutive past asset prices within the designated time window. In this work, we utilize asset price data derived from the S\&P 500 index, which provides historical stock prices for corporations listed on the index. Stock prices are inherently volatile, influenced by market activity, news, and other external factors. These short-term fluctuations often introduce noise that can adversely affect model performance. Therefore, in this paper, we preprocess the stock prices by  computing the finite difference between consecutive stock prices to capture the moving difference (returns). This is followed by performing a moving average, smoothing out these short-term fluctuations, revealing the underlying trends.

The value or movement of an asset at a time, denoted by $t,$ is represented by a random variable $X^t$ and we denote $X^t=x^t$ when it attains a value $x^t$. Thus, a context string of $T\geq 1$ numeric values is represented by a random vector $\boldsymbol{X}^{(T)}=(X^1,X^2,\hdots, X^{T})$. Then the task of a prediction model $f$ is to predict the value(s) of $\boldsymbol{X}^{(T+\tau)}\backslash \boldsymbol{X}^{(T)} := (X^{T+1},X^{T+2},\hdots, X^{T+\tau})$ as a probability distribution $f(\boldsymbol{X}^{(T+\tau) \backslash (T)}; \boldsymbol{X}^{(T)}) $ for some values of the future time $\tau\geq 0$. To be specific, given $\boldsymbol{X}^{(T)}=\boldsymbol{x}^{(T)},$ the task of the prediction model is to assign probabilities to future states $\boldsymbol{x}^{(T+\tau)}\backslash \boldsymbol{x}^{(T)} := (x^{T+1},x^{T+2},\hdots, x^{T+\tau}),$ the possible asset-price at the future times between $T+1$ and $T+\tau$, as $f(\boldsymbol{x}^{(T+\tau) \backslash T}; \boldsymbol{x}^{(T)}) $ and the efficiency of $f$ is determined by the closeness of a distance between $f(\boldsymbol{x}^{(T+\tau) \backslash T}; \boldsymbol{x}^{(T)})$ and the observed distribution over $ \boldsymbol{x}^{(T+\tau)}\backslash \boldsymbol{x}^{(T)}$ to zero. Thus the design of a prediction model $f(\boldsymbol{X}^{(T+\tau) \backslash (T)}; \boldsymbol{X}^{(T)}, \boldsymbol{\theta})$ with parameters $ \boldsymbol{\theta} $ is concerned with modeling $f$ such that the loss function $\mathcal{L}(f(\boldsymbol{x}^{(T+\tau) \backslash (T)}; \boldsymbol{x}^{(T)}), \boldsymbol{x}^{(T+\tau)}\backslash \boldsymbol{x}^{(T)})$, which estimates a notion of closeness between the probability distribution $f(\boldsymbol{X}^{(T+\tau) \backslash (T)}; \boldsymbol{X}^{(T)})$ and observed frequencies $\boldsymbol{X}^{(T+\tau)}\backslash \boldsymbol{X}^{(T)} $ is minimized for all or a collection of assets in a financial market. In most occasions dealing with time-series data models, the loss function is considered as the mean squared error (MSE), binary cross entropy or any other custom function specifically designed for the task. 

In our proposal of developing a QML model for multi-asset price prediction, it is customary to encode the context string data $\boldsymbol{x}^{(T)}$ into a quantum state which will be evolved under a unitary transformation. We define the $T$-qudit quantum state as
\begin{equation}
    \ket{\boldsymbol{x}^{(T)}}= \ket{x^1} \otimes \ket{x^2}\otimes \cdots \otimes \ket{x^{T}}
    \label{basis}
\end{equation}
to encode $\boldsymbol{x}^{(T)}$ after encoding the context data point $x^t$ into its corresponding qudit $\ket{x^t}$ for $1\leq t\leq T$. Here, $\otimes$ denotes the Kronecker product (also called tensor product) of two vectors. In this framework, the state of the qudit, $\ket{x^t}$, represents the quantized return of an asset at time $t$. The mapping of returns to the orthogonal states $\ket{0}, \ket{1}, \dots, \ket{d-1}$ is determined by dividing the range of possible returns into $d$ discrete intervals. Each interval corresponds to one of the orthogonal basis states of the qudit. For instance, let the minimum and maximum returns between $1\leq t\leq T$ be $x_{\text{min}}$ and $x_{\text{max}}$, then the price range is divided into $d$ equal intervals of length
\[\Delta x = \frac{x_{\max} - x_{\min}}{d-1}.\]
For any price $x^t$, its corresponding qudit state $\ket{x^t}$ is determined as:
\begin{equation} 
\ket{x^t} = \ket{i^t}, \quad i^t = \left\lfloor \frac{x^t - x_{\min}}{\Delta x} \right\rfloor, \quad i^t \in \{0, 1, \dots, d-1\}.
\label{encode}
\end{equation}
This mapping ensures that each basis state $\ket{i^t}$ represents a specific quantized interval of prices. Over time, as the asset price evolves, the state of the qudit $\ket{x^t} \in \mathbb{C}^{d}$ transitions between the basis states. These transitions can be modeled using a QNN $f(\boldsymbol{X}^{(T+\tau) / (T)}; \boldsymbol{X}^{(T)}, \boldsymbol{\theta}) $ with trainable parameters $\boldsymbol{\theta}$ designed to capture the stochastic behavior of returns and generate an approximation to the quantum state $ \sum_{\boldsymbol{x}^{(T+\tau) \backslash (T)} \in \{0,\ldots,d-1\}^{\tau}} \sqrt{f(\boldsymbol{x}^{(T+\tau) \backslash (T)}; \boldsymbol{x}^{(T)})} \ket{\boldsymbol{x}^{(T+\tau)}}$ such that direct measurement samples from the underlying distribution. Here, unlike the computational basis state $\ket{\boldsymbol{x}^{(T)}}$, the output vector $\ket{\boldsymbol{y}^{(T+\tau)}}$ is a superposition state of the form
\begin{equation}
    \ket{\boldsymbol{y}^{(T+\tau)}} = \sum_{\phi \in \{0,1, \dots, d-1\}^{T+\tau} }c(\phi)\ket{\phi},
    \label{output}
\end{equation}
where $\ket{\phi}=\ket{\phi^1}\otimes\ket{\phi^2}\dots\otimes\ket{\phi^{T+\tau}}$ is a $T+\tau$-qudit basis state with $\phi^{t}\in\{0,\dots,d-1\}$ and $c(\phi)\in \mathbb{C}$ such that $ \sum_{\phi} |c(\phi)|^2 = 1$. Now, the prediction is a density operator after taking the partial trace over the context: 
\begin{equation} 
\rho^{(T+\tau) \backslash (T)} = \text{Tr}_{1\ldots T} \ketbra{\boldsymbol{y}^{(T+\tau)}}{\boldsymbol{y}^{(T+\tau)}}. 
\end{equation}
Note that the prediction of future returns constitutes only the last $\tau$ qudits. However, in our situation, the QNN may make transformations to the context qudits ($\ket{\boldsymbol{x}^{(T)}}$), making it not suitable for reuse for repeated predictions. To circumvent that, we require the whole input and output vectors ($\ket{\boldsymbol{y}^{(T+\tau)}}$, $\ket{\boldsymbol{x}^{(T+\tau)}}$) to be involved in the loss function to make sure that the context qudits are unchanged. With this, we can approximate the prediction as a wavefunction:
\begin{equation}
    \ket{\boldsymbol{y}^{(T+\tau)}} \approx \sum_{\phi \in \{0,1, \dots, d-1\}^{\tau} }c(\phi)\ket{\boldsymbol{x}^{(T)}}\ket{\phi^{T+1}}\dots\ket{\phi^{T+\tau}},
    \label{y}
\end{equation}
where we note that the context state $\ket{\boldsymbol{x}^{(T+\tau)}}$ leads to a prediction $\sum_{\phi \in \{0,1, \dots, d-1\}^{\tau} }c(\phi)\ket{\phi^{T+1}}\dots\ket{\phi^{T+\tau}}$, which is a superposition of all possible outcomes with different probabilities. Each probable state $\ket{\phi^{T+t}}$ at a time $t$ ($T+1\leq t\leq T+\tau$) can be mapped to the numerical stock movement of an asset by the transformation
\begin{equation}
    \ket{\phi^{T+t}}\rightarrow x_{\min} + \phi^{T+t} \cdot \Delta x.
    \label{assign}
\end{equation}
These mapped values can then be used to calculate the expected value or movement of the stock prices by combining with the probability ($|c(\phi)|^2$) of each possible outcome. Measuring the last $\tau$ qudits over the computation basis states yields a sample from the underlying probability distribution over futures by the model: 
\begin{equation}
f(\phi^{T+1}, \ldots, \phi^{T+\tau}; \boldsymbol{x}^{(T)}, \boldsymbol{\theta}) = |c(\phi)|^2.
\end{equation}
Given $M$ measurement samples, we define the resulting distribution $f_{M}(\boldsymbol{X}^{(T+\tau) \backslash (T)}; \boldsymbol{x}^{(T)})$ based on the frequency of observing specific continuation strings $\boldsymbol{x}^{(T+\tau) \backslash (T)}$; in the high sample limit $ f_{M} \approx f $. We can also estimate the most probable future outcomes, which would be useful in repeated predictions. If we wish to predict a future state, $ \tau R + T $, we can use $R$ repeated application of the underlying QNN to generate a superposition over those outcomes. Fig.~\ref{fig:qnnflow} shows how a contextual QNN can be utilized for such time-series based predictions. \\

We summarize the time-series prediction model as
\begin{align*} 
\text{Context}:& \ket{\boldsymbol{x}^{(T)}}, \\ 
\text{Target}:& \ket{\boldsymbol{x}^{(T+\tau)}}, \\ 
\text{Model Prediction}:& \ket{\boldsymbol{y}^{(T+\tau)}}, \\ 
\text{Loss Function}:&     \;\mathcal{L}\Big(\ket{\boldsymbol{y}^{(T+\tau)}}, \ket{\boldsymbol{x}^{(T+\tau)}}\Big),
\end{align*}
In this paper, we are particularly interested in $\tau=1$ scenario along with binary quantization ($d=2$), for which the qudits are reduced to qubits such that $\ket{0}$ and $\ket{1}$ correspond to the negative and positive stock price movement respectively. Hereafter, we stick to the binary quantization $d=2$ throughout the paper unless otherwise stated. This choice provides a simple and interpretable mapping between price movements and quantum measurement outcomes, where each qubit directly represents an upward or downward movement in returns. Using binary encoding allows us to focus on the model’s ability to learn and reproduce the underlying conditional probability distributions, without additional complexity introduced by higher-dimensional encoding schemes. Moreover, binary quantization keeps the circuit depth and the number of qubits manageable, which is particularly important when studying scalability across multiple assets. By reducing representational overhead, we can isolate and analyze the learning dynamics of QMTL itself, establishing a clear baseline before extending the framework to multi-level quantization in later experiments. In addition, during training, we also consider the statistics of the contextual data including the contextual probability distribution $\mathcal{P}(\boldsymbol{X}^{(T)})$, the target probability distribution $\mathcal{P}(\boldsymbol{X}^{(T+1)})$, the conditional probability distribution $\mathcal{P}(X^{T+1}|\boldsymbol{X}^{(T)})$, and the total probability distribution $\mathcal{P}(X^{T+1},\boldsymbol{X}^{(T)})$. In particular, we train $ \boldsymbol{\theta} $ such that $ f(X^{T+1}; \boldsymbol{X^{(T)}}, \boldsymbol{\theta}) \approx \mathcal{P}(X^{T+1} | \boldsymbol{X}^{(T)}) $.

\subsection{QNN Framework}

Now we recall that a QNN architecture on an $n$-qubit register represents a parametrized unitary matrix $\hat{U}(\boldsymbol{\theta})$ of dimension $2^n$, which is defined by a sequence of parametrized quantum gates that produces an $n$-qubit output state $\hat{U}(\boldsymbol{\theta})\ket{\Psi}$ for any input state $\ket{\Psi}\in \mathbb{C}^{2^n},$ where $\boldsymbol{\theta}$ is the set of (real) parameters in the QNN and $\ket{\Psi}$ encodes a classical input data for the problem. The parameters in $\boldsymbol{\theta}$ can be learned and trained to produce a desired output state which is approximated by performing several quantum measurements to all or a subset of the qubits. Deciding the parametrized quantum circuit (PQC), also known as \textit{ansatz} which represents $\hat{U}(\boldsymbol{\theta})$ in a QNN model is one of the fundamental problems in QML applications. In this section, we will introduce the important components of a QNN.

\subsubsection{Loading Classical Data}

The first step in a QNN involves encoding classical data into quantum states. This is typically done using quantum feature maps~\cite{z}, where classical input data $\Psi \in \mathbb{C}^{\gamma}, 1\leq \gamma\leq 2^n$ is encoded into a quantum state $\ket{\Psi}$, where $2^n$ is its dimension. This state can be prepared by
\begin{equation}
    \ket{\Psi} = \hat{U}_{F}(\Psi)\ket{0}^{\otimes n}, 
    \label{map}
\end{equation}
where $\hat{U}_{F}(\Psi)$ is a quantum feature map circuit that depends on the classical data $\Psi$, and $\ket{0}^{\otimes n}$ represents the initial quantum state. A feature map consists of a set of controlled rotation gates, parameterized by the contents in the classical register $\Psi$. Different feature maps can be employed depending on the application to project the classical data on the quantum state space. Some of the common feature maps used are the first and second-order Pauli-Z evolution circuits \cite{evolution}. 

\subsubsection{Parametric Quantum Circuits}

\begin{figure}[t]
\centering
\includesvg[width=0.9\linewidth]{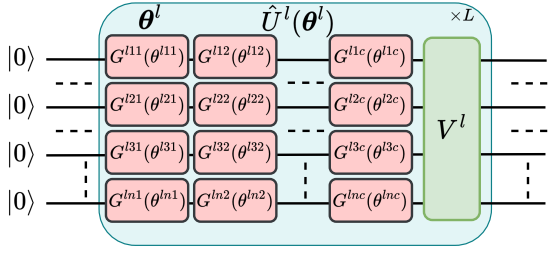}
\caption{\textbf{Parameterized Quantum Circuit}. Block diagram of the layered parametric quantum circuit showing various blocks in the $l^{th}$ layer such as fixed unitary and parametric rotation gates.} 
\label{blocks}
\end{figure}

Once the classical data is encoded into a quantum state, it is processed by a PQC. The goal of the training process is to optimize the parameters $\boldsymbol{\theta}$ such that the PQC outputs a quantum state that corresponds to accurate predictions for the learning task. In this paper, inspired from the traditional layered neural networks, we focus on PQCs with $L$ repeated layers, composed on fixed and parameterized gates such that the unitary represented by the circuit at layer $l\in[1,L]$ is $\hat{U}^l(\boldsymbol{\theta}^l) = V^{l}\prod_{j} G^{lij}(\theta^{lij})$ as shown in Fig.~\ref{blocks}, where $V^{l}$ is a fixed unitary at layer $l$ (which could be identity or a sequence of CNOT gates) and $G^{lij}(\theta^{lij})$ denote a single-qubit rotation gate acting at layer $l$, at position $(i,j)$. Here, $i\in[1,n]$ and $j\in[1,c]$, with $i$ and $j$ corresponding to the row and column indices of the quantum circuit, respectively as shown in Fig.~\ref{blocks}. In this notation, $\boldsymbol{\theta} = \bigoplus_{l=1}^L\boldsymbol{\theta}^l$, where $\boldsymbol{\theta}^l = \{\theta^{l11}, \theta^{l12}, \hdots, \theta^{lnc}\}$ such that $m(=Lnc)$ is the total number of parameters in the circuit, $n(=m/Lc)$ is the number of of qubits and $c(=m/Ln)$ is the number of sub-layers in each layer. This representation will be explored in detail in later sections. Each parameter $\theta^{lij} \in \boldsymbol{\theta} $ can correspond to the angles of single qubit rotation gates such as $R_{X}(\theta^{lij}), R_{Y}(\theta^{lij}), R_{Z}(\theta^{lij})$. In the proposed QML model in this paper, along with the context quantum state $\ket{\boldsymbol{x}^{(T)}}$, we need ancilla qubits to extract information from the output state of a QNN for the learning task. The ancilla qubit states control the application of certain quantum gates on the main quantum register to obtain a desired output state. Assuming that there are $\tau$ ancilla qubits, the QNN represents a unitary matrix $\hat{U}(\boldsymbol{\theta})$ with dimension $2^{T+\tau} \times 2^{T+\tau}$, resulting in the number of qubits $n=T+\tau$. Setting the initial state of the ancilla register as $\ket{0}^{\otimes \tau},$ the output state is given by 
\begin{equation}
\ket{\boldsymbol{y}^{(T+\tau)}} = \hat{U}(\boldsymbol{\theta})\big(\ket{\boldsymbol{x}^{(T)}}\otimes\ket{0}^{\otimes \tau}\big),
\label{pqc}
\end{equation}
where $\hat{U}(\boldsymbol{\theta})=\prod_{l}\hat{U}^l(\boldsymbol{\theta}^l)$ is the unitary operator corresponding to the parametric quantum circuit. In addition to rotation gates, our circuit may include fixed gates, such as CNOT gates, that help spread entanglement in the system. 

\subsubsection{Training}
Classical neural networks are primarily trained by backpropagation, which is not directly possible in QNNs. Quantum states collapse upon measurement, meaning the quantum information is destroyed. Backpropagation relies on preserving intermediate computations (like activations) during the forward pass for use in the backward pass. In quantum systems, measurements required to extract information disrupt the state, preventing reuse. In addition, the no-cloning theorem prohibits copying quantum states for reuse, further complicating backpropagation. Consequently, QNNs have adopted to other techniques for gradient computation such as parameter shift \cite{parameter} and simultaneous perturbation stochastic approximation (SPSA) \cite{spsa}. Unlike backpropagation, which requires explicit differentiation through each layer, these methods compute gradients by evaluating the quantum circuit at slightly shifted parameter values. It works based on the fact that the output of quantum circuits often depends on the parameters through trigonometric functions (such as sine and cosine), enabling exact gradient computation.

If $\hat{B}$ be the observable that we would like to measure, then the expectation value of the output state of the PQC with respect to $\hat{B}$ is $\langle \hat{B} \rangle_{\boldsymbol{\theta}}=\big\langle \boldsymbol{x}^{(T+\tau)} \big|\hat{U}^{\dagger} (\boldsymbol{\theta})\hat{B} \hat{U}(\boldsymbol{\theta}) |\boldsymbol{x}^{(T+\tau)} \rangle$. With this, the gradient update rules are given by\cite{parameter}\cite{spsa}:\\
\begin{enumerate}
    \item \textit{Parameter-Shift}\\
    \begin{equation}
    \frac{\partial}{\partial \theta^{lij}} \langle \hat{B} \rangle_{\boldsymbol{\theta}} = \frac{1}{2} \left( \langle \hat{B} \rangle_{\theta^{lij} + \frac{\pi}{2}} - \langle \hat{B} \rangle_{\theta^{lij} - \frac{\pi}{2}} \right),
    \label{shift} \end{equation}
    \item \textit{SPSA}
    \begin{equation}
    \frac{\partial}{\partial \theta^{lij}} \langle \hat{B} \rangle_{\boldsymbol{\theta}} = \frac{1}{2\delta \alpha^{lij}} \left( \langle \hat{B} \rangle_{\boldsymbol{\theta} + \delta\boldsymbol{\alpha}} - \langle \hat{B} \rangle_{\boldsymbol{\theta} -  \delta\boldsymbol{\alpha}} \right),
    \label{spsa} \end{equation}
\end{enumerate}
where $\delta$ is hyperparameter, typically between $0$ and $1$ and $\boldsymbol{\alpha}$ is a stochastic perturbation vector with dimensions same as $\boldsymbol{\theta}$. The vector $\boldsymbol{\alpha}$ is sampled from a zero mean distribution such that $\alpha^{lij}\in\{-1,1\}$. The primary difference between the two approaches is in parameter updates: parameter-shift updates one parameter at a time, requiring $2m$ evaluations for $m$ parameters, while SPSA updates all parameters simultaneously with just two evaluations per iteration.

\subsubsection{Loss Functions}

Different loss functions are employed depending on the nature of the QNN and the gradient update rule. For instance, mean squared error (MSE) is commonly used in regression tasks due to its smooth gradients and simplicity in optimization. However, for classification tasks, cross-entropy loss might be preferred as it better captures the probabilistic nature of the output. Therefore, we introduce different loss functions that are relevant to our paper.

Mean squared error loss is the most common loss function used for optimization tasks and in our context, it is defined as:
\begin{equation}
    \mathcal{L}^{m}(\boldsymbol{\theta}, \boldsymbol{x}^{(T+\tau)}) = | f_{M}(x^{(T+\tau) \backslash (T)} ; x^{(T)}) - \delta(\boldsymbol{x}^{(T+\tau)}) |^2.
\label{mse_eq}
\end{equation}
Note that for the MSE loss, full state measurement is required i.e., all the qubits must be measured and the returns have to retrieved from the output state vector as per Eq.~\eqref{assign}. To circumvent this, we use the fidelity loss or the SWAP test~\cite{swap} to compare the output state $\ket{\boldsymbol{y}^{(T+\tau)}}$ with the target state $\ket{\boldsymbol{x}^{(T+\tau)}}$. An ancilla qubit is prepared in the state $ \frac{1}{\sqrt{2}} \left( \ket{0} + \ket{1} \right)$, whose logical state is then used to control SWAP the wavefunction (e.g. qubit by qubit). Thereafter, a Hadamard gate is applied on the ancilla, leading to a phase kick back, and then the ancilla is measured in the computational (Pauli-Z) basis. 
\begin{figure}[t]
\includesvg[width=0.9\linewidth]{imgs/qmtl3.svg}
\caption{\textbf{SWAP Test.} A diagram of the SWAP test, which measures the fidelity loss between two wavefunction states $\ket{\boldsymbol{x}^{(T+\tau)}}$ and $\ket{\boldsymbol{y}^{(T+\tau)}}$.} 
\label{swap}
\end{figure}

The probability of obtaining $\ket{0}$ after measurement of the ancilla qubit is proportional to the fidelity $\big|\big\langle \boldsymbol{y}^{(T+\tau)} \ket{{\boldsymbol{x}}^{(T+\tau)}}\big|^2$ and is given by:
\begin{equation}
    \mathcal{P}(0) = \frac{1 + \big|\big\langle \boldsymbol{y}^{(T+\tau)} \ket{{\boldsymbol{x}}^{(T+\tau)}}\big|^2}{2}
    \label{prob_swap}
\end{equation}
and the corresponding fidelity loss is:
\begin{equation}
    \mathcal{L}^{f}(\boldsymbol{\theta}, \boldsymbol{x}^{(T+\tau)}) = 1 - \mathcal{P}(0) =  \frac{1 - \big|\big\langle \boldsymbol{y}^{(T+\tau)} \ket{{\boldsymbol{x}}^{(T+\tau)}}\big|^2}{2}
    \label{loss_swap}
\end{equation}

\section{Loading Classical Data}

\begin{figure}[t]
\centering
\includesvg[width=1.1\linewidth]{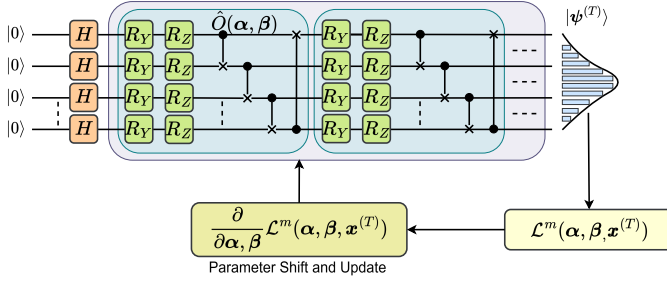}
\caption{\textbf{Loading a distribution onto a hardware efficient ansatz.} The circuit inside the blue box ($\hat{O}(\boldsymbol{\alpha},\boldsymbol{\beta})$) is applied sequentially for a sufficient number of iterations followed by an MSE loss with the SPSA update rule.} 
\label{loading}
\end{figure}

Contextual data $\boldsymbol{x}^{(T)}$ can be encoded onto a quantum state using feature maps \cite{evolution}, as described in Eq.~\eqref{map}. However, feature maps are computationally intensive and slow to train due to the convoluted nature of $\hat{U}_f$. Moreover, training requires encoding a different contextual input during each iteration, further increasing the complexity. To address this challenge, we propose loading the entire contextual distribution onto the quantum state using a hardware-efficient ansatz \cite{hec}. By encoding the data only once, our approach achieves significantly faster training. In this section, we discuss the procedure for encoding the contextual probability distribution $\mathcal{P}(\boldsymbol{X}^{(T)})$, where $\boldsymbol{X}^{(T)}$ represents the contextual time-series data of an asset, onto a quantum state using the hardware efficient ansatz architecture. 

Once the data has been preprocessed, we construct the Hilbert space as per Eq ~\eqref{encode} to obtain a basis for the $T$-qubit representation of the returns, where the quantization levels for the returns are also chosen. Now, we need to represent the stochastic behavior of the asset in the space defined by the basis vectors such that the information can be used to train the QNN model and make reasonable predictions. To this end, we normalize the preprocessed data and compute its histogram to obtain the contextual probability distribution of the quantized returns as $\mathcal{P}({\boldsymbol{X}}^{(T)}=(i^1,\dots,i^T))$, where $i^t$ denotes the quantized return at time $t$. Thereafter, we load the contextual distribution onto the $T$-qubit state as follows 
\begin{equation}
    \ket{\boldsymbol{\psi}^{(T)}} = \sum_{i^1,\dots,i^T \in \{0,1\} } \sqrt{\mathcal{P}(i^1,\dots,i^T)} \ket{i^1}\otimes\ket{i^2}\dots\otimes\ket{i^{T}}
    \label{psi}
\end{equation}
where $\mathcal{P}(i^1,\dots,i^T)$ is the value of the quantized distribution over the corresponding basis state $\ket{i^1}\otimes\ket{i^2}\dots\otimes\ket{i^{T}}$. Note that, using the definition of $\boldsymbol{x}^{(T)}$ and equations ~\eqref{basis} and ~\eqref{encode}, we can rewrite Eq.~\eqref{psi} as
\begin{equation}
    \ket{\boldsymbol{\psi}^{(T)}} = \sum_{\boldsymbol{x}^{(T)}\in \{0,1\}^{T}} \sqrt{\mathcal{P}(\boldsymbol{x}^{(T)})} \ket{\boldsymbol{x}^{(T)}}
    \label{load}
\end{equation}

Eq.~\eqref{load} can be achieved using the Grover-Rudolph technique for state preparation \cite{grover-rudolph}. However, due to its higher computational complexity, we adopt a simpler machine learning-based approach for state preparation. This method utilizes a hardware-efficient ansatz \cite{hec} combined with a mean squared error (MSE) loss function (Eq.~\eqref{mse_eq}), where the parameters of the rotation gates are iteratively updated in a loop using the SPSA rule to minimize the loss function, as illustrated in Fig.\ref{loading}. This circuit is chosen due to its hardware-efficient architecture \cite{hec_proof}, which constitutes a repeated layers of $R_{Y}$, $R_{Z}$, and $CNOT$ gates to perform rotations and entanglement. Although several loss functions exist for comparing distributions, we choose MSE due to its smooth gradients and computational efficiency, enabling faster optimization. For each layer, the corresponding unitary transformation looks like:
\begin{equation}
    \hat{O}(\boldsymbol{\alpha}, \boldsymbol{\beta}) = \prod_{j=0}^{T-1} CNOT_{j,(j+1)\%T}\bigotimes_{j=0}^{T-1} R_Z(\beta^j) \bigotimes_{j=0}^{T-1} R_Y(\alpha^j)
\end{equation}
where $CNOT_{j,(j+1)\%T}$ acts on qubits $j$ (control) and $(j+1)\%T$ (target) and $R_{Y}(\alpha^i)$, $R_{Z}(\beta^i)$ are $2\times 2$ unitary matrices acting on the $j^{th}$ qubit. To enhance the accuracy of loading the distribution, multiple such transformations are applied iteratively, as illustrated in the figure. This approach improves the learnability of the circuit, thereby increasing the fidelity of the loaded distribution. Note that the parameters in the circuit $\{\{\alpha^i\}\cup\{\beta^i\}\}$ are optimized through the training process to accurately load the contextual distribution.

\begin{figure}[t]
\centering
\includesvg[width=\linewidth]{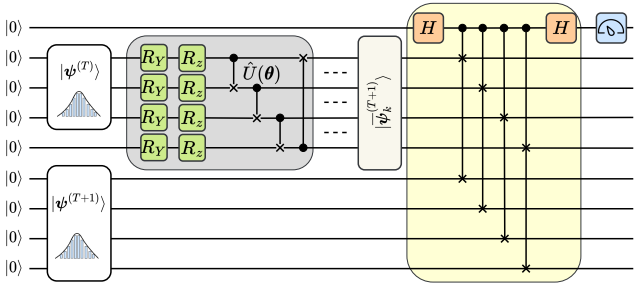}
\caption{\textbf{Quantum Batch Learning}. A diagram showing the learning procedure of our proposed quantum batch gradient update for a context of $T$. The top most qubit is an ancilla qubit. For a batch, a distribution over inputs is loaded succeeding qubits (the input qubits) and the following qubit(s) is for the output. The joint distribution of the inputs and outputs for the batch is loaded on the subsequent qubits. The circuit inside the Grey box ($\hat{U}(\boldsymbol{\theta})$) is applied sequentially for a required number of iterations, is a contextual quantum neural network to prepare an approximate of the loaded joint distribution. A SWAP test is then used to take the fidelity loss between the two distributions.} 
\label{stl_arc}
\end{figure}

\section{Quantum Single-Task Learning}

Now that we have encoded the contextual distribution into the quantum states, we can move on to discussing predictions based on the given context. Before diving into quantum multi-task learning for predicting multiple asset prices, we'll first present a complete case of using quantum circuits for predicting the price of a single stock, which we will refer to as quantum single-task learning (QSTL). 

Unlike previous approaches~\cite{qgan,qwgan}, which rely only on using the entire historical data to make predictions, we incorporate contextual information as well to forecast future outcomes, as shown in Fig.~\ref{stl_arc}. This method offers the advantage of adapting to the constantly changing stock market, whereas past methods may become less relevant due to outdated data. In our approach, the contextual distribution is encoded onto the wavefunction $\ket{\boldsymbol{\psi}^{(T)}}$, which, along with a prediction qubits, forms the input of the quantum circuit: $\ket{\boldsymbol{\psi}^{(T)}}\otimes \ket{0}^{\otimes \tau}$. Hereafter, we will stick to $\tau=1$ and binary quantization unless otherwise stated. We also load the target distribution $\ket{\boldsymbol{\psi}^{(T+1)}}$ to the last $T+1$ qubits, which serves as an input to the SWAP test. The layered PQC forms the bulk of the circuit followed by the SWAP test between the predicted distribution and target distribution (as shown in Fig.\ref{stl_arc}). Finally, measurement is done on the ancillary qubit, obtaining the fidelity loss, which guides the training process through SPSA gradient update rule.

A unitary transformation $\hat{U}(\boldsymbol{\theta})$, where $\boldsymbol{\theta}$ is the trainable parameters, enables the circuit to learn the conditional probability distribution represented by $\mathcal{P}(X^{T+1}|\boldsymbol{x}^{(T)},\boldsymbol{\theta})$. This probability distribution serves as the desired output state of the PQC at the end of the training process, allowing for the prediction of future returns via measurement. Learning conditional probability distributions is crucial for time-series prediction as they capture the dependency between past context and future outcomes. In contrast, marginal distributions are unsuitable for time-series prediction because they lack the ability to incorporate temporal dependencies and contextual information.

\subsection{Quantum Batch Gradient Update}

Quantum data processing has been explored for Machine Learning due to the inherent memory compression associated. For example Ref.~\cite{harrow2020} utilizes access in superposition to models for quantum speedup in tasks like k-means clustering, while Ref.~\cite{svm} achieves quantum speedup for Support Vector Machines by learning over a superposition of data points. In the context of QNNs, Ref.~\cite{wu2025} recently considered a construction for parallel quantum batches in which a reduced density matrix is constructed over the ancilla, output, and label space. In comparison, our approach loads the entire contextual distribution at once according to Eq.~\eqref{load}, resulting in a straight forward loss function that leads to higher quality gradients. Then our QNN generates an \textit{approximate} representation of the continuation for \textit{each} context in superposition. This leverages the linearity of quantum circuits, allowing a superposition of all possible inputs to train the circuit without breaking the correspondence between the respective inputs and outputs. This correspondence reduces the multi-step stochastic gradient descent to a single step. 

For example, consider Eq.~\eqref{pqc}, when the input of the PQC is a single context $\ket{\boldsymbol{x}^{(T)}}$ and the forward pass of this input through the PQC is $\ket{\boldsymbol{y}^{(T+1)}} = \hat{U}(\boldsymbol{\theta})\big(\ket{\boldsymbol{x}^{(T)}}\otimes\ket{0}\big)$ as shown in Fig.~\ref{stl_arc}.
Let the gradient update for $\boldsymbol{\theta}$, obtained from SPSA, be denoted as $g(\boldsymbol{\theta}, \boldsymbol{x}^{(T+1)})$, with entry $lij$:
\begin{align}
    g(\theta^{lij}, \boldsymbol{x}^{(T+1)}) &= \frac{\partial}{\partial \theta^{lij}}\mathcal{L}^{f}(\boldsymbol{\theta}, \boldsymbol{x}^{(T+\tau)}) \nonumber\\
    &= \frac{\partial}{\partial \theta^{lij}} \Big(\frac{1 - \big|\big\langle \boldsymbol{y}^{(T+\tau)} \ket{{\boldsymbol{x}}^{(T+\tau)}}\big|^2}{2}\Big)\ \nonumber\\
    &= -\frac{1}{2}\frac{\partial}{\partial \theta^{lij}} \bra{\boldsymbol{y}^{(T+\tau)}} \hat{B}\ket{{\boldsymbol{y}}^{(T+\tau)}} \nonumber\\
    &= -\frac{1}{4\delta \alpha^{lij}} \left( \langle \hat{B} \rangle_{\boldsymbol{\theta} + \delta\boldsymbol{\alpha}} - \langle \hat{B} \rangle_{\boldsymbol{\theta} -  \delta\boldsymbol{\alpha}} \right),
\end{align}
where the last step is obtained by using Eq.~\eqref{spsa} with $\hat{B} = \ketbra{\boldsymbol{x}^{(T+\tau)}}{\boldsymbol{x}^{(T+\tau)}}$. Therefore, the gradient update of $\boldsymbol{\theta}$ becomes
\begin{equation}
    \Rightarrow \boldsymbol{\theta} := \boldsymbol{\theta}-  \beta \, g(\boldsymbol{\theta},\boldsymbol{x}^{(T+1)}),
    \label{bpass1}
\end{equation}
where $\beta$ is the learning rate. Similarly, if the input to the PQC is a distribution of basis states (superposition of all possible contexts) such as $\ket{\boldsymbol{\psi}^{(T)}}$ from Eq.~\eqref{load}, then the forward pass through the PQC is given by:
\begin{align*}
\ket{\overline{\boldsymbol{\psi}}^{(T+1)}} &= \hat{U}(\boldsymbol{\theta}) \Big(\sum_{\boldsymbol{x}^{(T)}\in \{0,1\}^{T}} \sqrt{\mathcal{P}(\boldsymbol{x}^{(T)})}\ket{\boldsymbol{x}^{(T)}}\otimes\ket{0}\Big) \\
&= \sum_{\boldsymbol{x}^{(T)}\in \{0,1\}^{T}} \sqrt{\mathcal{P}(\boldsymbol{x}^{(T)})} \hat{U}(\boldsymbol{\theta}) \Big(\ket{\boldsymbol{x}^{(T)}}\otimes\ket{0}\Big)\\
&= \sum_{\boldsymbol{x}^{(T)}\in \{0,1\}^{T}} \sqrt{\mathcal{P}(\boldsymbol{x}^{(T)})} \ket{\boldsymbol{y}^{(T+1)}}, \\
\end{align*}
where $\ket{\overline{\boldsymbol{\psi}}^{(T+1)}}$ is the total distribution learned by the circuit $\hat{U}(\boldsymbol{\theta})$ over the contextual distribution $\ket{\boldsymbol{\psi}^{(T)}}$. Given that the gradient update corresponding to the output $\ket{\boldsymbol{y}^{(T+1)}}$ is $g(\boldsymbol{\theta},\boldsymbol{x}^{(T+1)})$ (from Eq.{~\ref{bpass1}}), then the gradient update for the output $\ket{\overline{\boldsymbol{\psi}}^{(T+1)}}$ can be calculated from the derivative over summation rule as
\begin{equation}
    \Rightarrow \boldsymbol{\theta} := \boldsymbol{\theta}-\beta\sum_{\boldsymbol{x}^{(T)}\in \{0,1\}^{T}} \mathcal{P}(\boldsymbol{x}^{(T)}) \, g(\boldsymbol{\theta},\boldsymbol{x}^{(T+1)}).
    \label{bpass2}
\end{equation}
This expression (Eq.{~\ref{bpass2}}) effectively corresponds to applying stochastic gradient descent (SGD) across all input context samples from the dataset, achieving a single gradient update after processing the entire dataset. Notably, this result is obtained in one step due to the inherent linearity of quantum mechanics, facilitating the quantum circuit to train on large batches. Using the QBGU training process, the model $\hat{U}(\boldsymbol{\theta})$ in Fig.~\ref{stl_arc} is trained to learn the conditional probability distribution $\mathcal{P}(x^{T+1}|\boldsymbol{x}^{(T)},\boldsymbol{\theta})$, which essentially maps each input $\ket{\boldsymbol{x}^{(T)}}$ to its corresponding output $\ket{\boldsymbol{y}^{(T+1)}}$ at the time of inference. As shown in Fig.~\ref{stl_arc}, we preload both the contextual distribution $\ket{\boldsymbol{\psi}^{(T)}}$ and the target distribution $\ket{\boldsymbol{\psi}^{(T+1)}}$, and the SWAP test then measures the distance between the estimated and the original target distributions, guiding the training process.

\begin{figure}[t]
\centering
\includesvg[width=\linewidth]{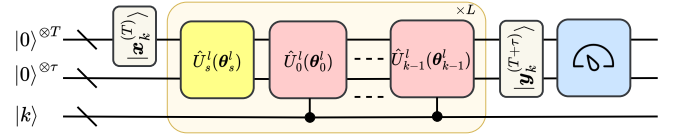}
\caption{\textbf{Share-and-specify Ansatz.} A diagram of our Quantum Multi-Task Learning Architecture showing various components for the QNN. A single asset state is loaded by setting the label $\ket{k}$ and the inference-time context $\boldsymbol{x}^{(T)}$ is loaded over qubits $\ket{0}^{\otimes T+\tau}$. The input can then be processed through the parameterized circuit, composed of $L$ layers of the share-and-specify ansatz, to define a state $\ket{\boldsymbol{y}^{(T)}} $ that can be utilized for a downstream task. For example, as depicted in the figure, measurement can be used to sample possible continuations $ x^{\tau} $.} 
\label{arch}
\end{figure}

\section{Quantum Multi-Task Learning}
Multi-task learning (MTL) \cite{mtl} is a machine learning approach where multiple tasks are learned at the same time, allowing the model to share information between them. It uses shared parameters to capture common patterns across all tasks, while task-specific parameters focus on unique aspects of each task. In financial time series prediction, MTL can be used to predict the prices or trends of multiple assets together. Shared parameters can represent factors that affect the entire market, such as economic indicators, while task-specific parameters account for unique characteristics of each asset, like individual volatility or trading patterns. This helps the model make better predictions by learning both shared and asset-specific information. In this section, we extend these ideas to QNNs by introducing the quantum multi-task learning (QMTL).

\begin{figure*}[t]
\centering
\includesvg[width=\linewidth]{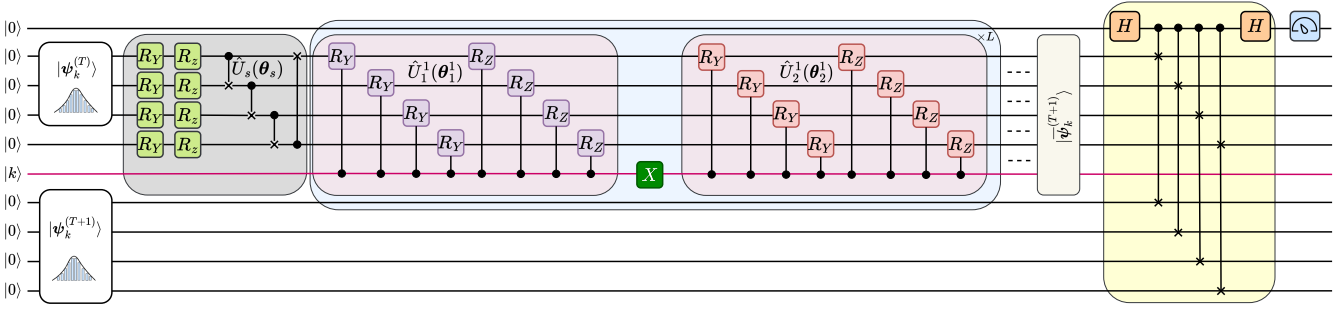}
\caption{\textbf{Quantum Multi-Task Learning Architecture for $T=3$ and $K=2$.} The circuits inside the Grey ($\hat{U}_s(\boldsymbol{\theta_s})$) and Pink boxes ($\hat{U}_1^1(\boldsymbol{\theta}_1^1)$, $\hat{U}_2^1(\boldsymbol{\theta}_2^1)$) are applied sequentially for a required number of iterations followed by the SWAP test.} 
\label{mtl_arc}
\end{figure*}
\subsection{Circuit Architecture}
In order to incorporate MTL in a QNN framework, we introduce the \textit{share-and-specify} ansatz (Fig.~\ref{arch}) which breaks each layer of the PQC into a block of universal gates (\textit{shared ansatz}), followed by a block of asset specific label-controlled gates acting based on the state of the label registers (\textit{specify ansatz}). The share ansatz for each layer can be embodied by the same gates used in the single-asset task, leading to identity operations applied to the $\log K$ label qubits, 
\begin{equation}
\hat{U}^l_s(\boldsymbol{\theta}^l_s) = \Big(V^{l}_s\otimes\mathbbm{1}_{K}\Big) \prod_{i,j}\Big(G_s^{lij}(\theta^{lij}_s) \otimes\mathbbm{1}_{K}\Big),
\label{s}
\end{equation}
where $\mathbbm{1}_{K}$ is the identity operator with dimension $K\times K$. The label qubits, collectively represented by the qudit $\ket{k} = \ket{k_1}\otimes\ket{k_2}\dots\otimes\ket{k_{\log K}}$ with $k_j\in\{0,1\}$, are used to distinguish the assets. Each asset is assigned a unique label $k\in[1,K]$, ensuring that the quantum operations are applied selectively to the corresponding asset based on its label. These label qubits are then used to form the task-specific unitary operator for asset $k$ as 
\begin{equation}
\hat{U}^l_k(\boldsymbol{\theta}^l_k) = \Big(V^{l}_k\otimes\mathbbm{1}_{K}\Big) \prod_{i,j}\Big(G_k^{lij}(\theta^{lij}_k) \otimes\mathbbm{1}_{K}\Big),
    \label{k}
\end{equation} 
resulting in the specify ansatz (with control) for asset $k$ as 
\begin{align}
    \hat{U}^l_{ck}(\boldsymbol{\theta}^l_{ck}) &= (V^{l}_k \otimes \ketbra{k}{k} + \mathbbm{1}_{2^{T+\tau}} \otimes \ketbra{k^\perp}{k^\perp}) \nonumber \\ 
    &\phantom{= } \prod_{i,j}\Big(G_k^{lij}(\theta^{lij}_{k}) \otimes \ketbra{k}{k} + \mathbbm{1}_{2^{T+\tau}} \otimes \ketbra{k^\perp}{k^\perp} \Big),
    \label{ck}
\end{align}
where $\boldsymbol{\theta}^l_{ck}=\boldsymbol{\theta}^l_{k}$, but the subscript $c$ is added for notational consistency and $\ketbra{k^\perp}{k^\perp} = \mathbbm{1}_{K} - \ketbra{k}{k}$ is a projection onto the orthogonal space of $\ketbra{k}{k}$, such that $\bigotimes_{i}G_k^{lij}(\theta^{lij}_k)$ is only applied when the label qudit is $\ket{k}$. In particular, if $ G_{k}^{lij} \in \mathbbm{1}_{2^{i-1}} \otimes \{ \hat{R}_X(\theta^{lij}_k), \hat{R}_Y(\theta^{lij}_k), \hat{R}_Z(\theta^{lij}_k) \} \otimes \mathbbm{1}_{2^{n-i}}$, then $\big(G_k^{lij}(\theta^{lij}_k) \otimes \ketbra{k}{k} + \mathbbm{1}_{2^{T+\tau}} \otimes \ketbra{k^\perp}{k^\perp} \big)$ is a control rotation based on label $\ket{k}$. We can then define the specify ansatz for one layer as $ \hat{U}_{c}^{l}(\boldsymbol{\theta}_c^l) = \prod_{k} \hat{U}_{ck}^{l}(\boldsymbol{\theta}^{l}_{ck}) $ and the entire PQC as 
\begin{equation}
    \hat{U}(\boldsymbol{\theta}) = \prod_{l=1}^{L} \hat{U}_{c}^{l}(\boldsymbol{\theta}_c^{l}) \hat{U}_{s}^{l}(\boldsymbol{\theta}_{s}^l),
    \label{u}
\end{equation}
where the dimension of $\hat{U}(\boldsymbol{\theta})$ is $2^{T+\tau}K \times 2^{T+\tau}K$. Note that, $\boldsymbol{\theta} = \bigoplus_{l=1}^{L} \boldsymbol{\theta}_s^l \oplus\boldsymbol{\theta}_c^l $. For each asset $k$, the contextual price data of size $T$, represented as $\ket{\boldsymbol{x}^{(T)}_{k}}$, undergoes the transformation determined by the label qudit and the task-specific ansatz:

\begin{align*}
\ket{{\boldsymbol{y}}^{(T+1)}_{k}} &= \hat{U}(\boldsymbol{\theta})\big(\ket{\boldsymbol{x}^{(T)}_{k}}\otimes\ket{0}^{\otimes \tau} \otimes \ket{k}\big)\\
&= \prod_{l} \hat{U}_{c}^{l}(\boldsymbol{\theta}^{l}) \Big(\hat{U}_{s}^{l}(\boldsymbol{\theta}_{s}^l) \big(\ket{\boldsymbol{x}^{(T)}_{k}}\otimes\ket{0}^{\otimes \tau}\big)\otimes \ket{k}\Big)\\
 &=\prod_{l} \hat{U}_{k}^{l}(\boldsymbol{\theta}_{k}^l)\hat{U}_s^{l}(\boldsymbol{\theta}_{s}^l)\big(\ket{\boldsymbol{x}^{(T)}_{k}}\otimes\ket{0}^{\otimes \tau}\big)\otimes \ket{k}.
 \label{mtl}
\end{align*}
Intuitively, the share ansatz layer $\hat{U}_{s}^{l}$ helps facilitate learning across assets while the specify ansatz layer $\hat{U}_{k}^{l}$ allows focus on potential peculiarities of an individual asset. This entire architecture is demonstrated in Fig.~\ref{arch}.

\begin{figure*}[t]
\centering
\includesvg[width=\linewidth]{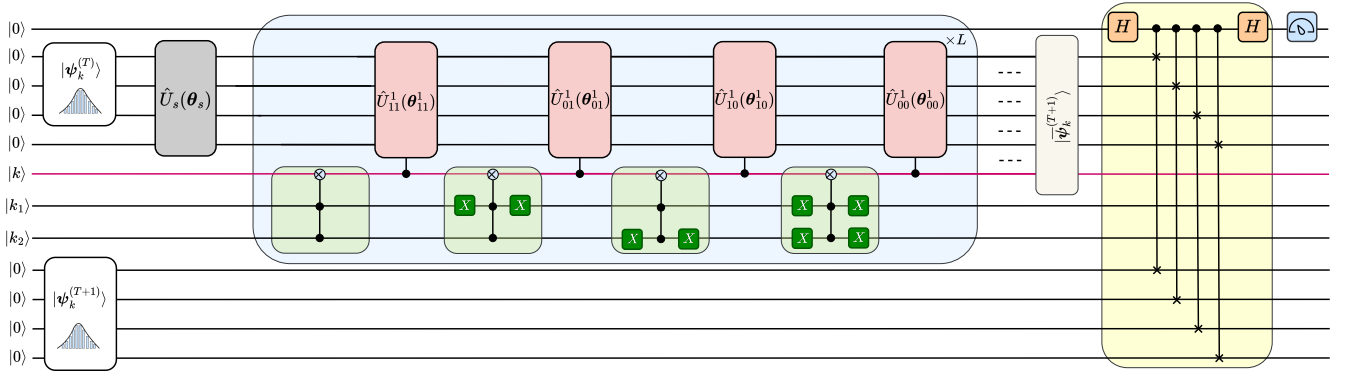}
\caption{\textbf{Quantum Multi-Task Learning Architecture for $T=3$ and $K=4$.} The circuits inside the Grey ($\hat{U}_s(\boldsymbol{\theta}_s)$) and Pink boxes ($\hat{U}_{11}^1(\boldsymbol{\theta}_{11}^1)$, $\hat{U}_{01}^1(\boldsymbol{\theta}_{01}^1)$, $\hat{U}^1_{10}(\boldsymbol{\theta}_{10}^1)$,  $\hat{U}^1_{00}(\boldsymbol{\theta}_{00}^1)$) are applied sequentially for a required number of iterations followed by the SWAP test.} 
\label{4mtl_arc}
\end{figure*}

\subsubsection{Two-Assets Case}

Extending the previous framework to handle two assets, the forward pass equations for a set of inputs $\ket{\boldsymbol{x}_{1}^{(T)}}$, $\ket{\boldsymbol{x}_{2}^{(T)}}$ for two different assets at layer $l$ are given as
\[\ket{\boldsymbol{y}_{1}^{(T+1)}} = \hat{U}_1^l(\boldsymbol{\theta}_1^l)\hat{U}_s^l(\boldsymbol{\theta}_s^l) \big(\ket{\boldsymbol{x}_{1}^{(T)}}\otimes\ket{0}\otimes\ket{0}\big)\] 
\[\ket{\boldsymbol{y}_{2}^{(T+1)}} = \hat{U}_2^l(\boldsymbol{\theta}_2^l)\hat{U}_s^l(\boldsymbol{\theta}_s^l) \big(\ket{\boldsymbol{x}_{2}^{(T)}}\otimes\ket{0}\otimes\ket{1}\big)\] 

The subscripts $1$ and $2$ denote first and second assets, respectively, with the last qubit serving as the control to switch between assets. As illustrated in Fig.~\ref{mtl_arc}, the unitary operator $\hat{U}_s(\boldsymbol{\theta}_s^l)$ is shared across all assets, while the operators $\hat{U}_1^l(\boldsymbol{\theta}_1^l)$ and $\hat{U}_2(\boldsymbol{\theta}_2^l)$ are task-specific trainable PQCs. The simplest way to switch between tasks is by applying an $X$ gate, as illustrated in Fig.~\ref{mtl_arc}. 

\subsubsection{Four-Assets case}
For scenarios involving more than two assets or tasks, constructing the control qudit using only single-qubit gates and CNOT gates leads to a significant increase in the number of label qubits. To address this, our approach incorporates Toffoli gates, which enables us to minimize the number of label qubits to $\log K+1$. For instance, with four assets, we designed a control circuit employing three qubits, Toffoli gates, and $X$ gates, as depicted in Fig.~\ref{4mtl_arc}. This configuration ensures that exactly one of the transformations $\hat{U}_k^l(\boldsymbol{\theta}_k^l)$ is active for each unique combination of $\ket{k_1} \otimes \ket{k_2}$, corresponding to a specific asset. The task-specific PQCs and their associated controls on the qubits $(\ket{k} \otimes \ket{k_1} \otimes \ket{k_2})$ can be expressed as:
\begin{align}
    \hat{U}_{11}^l(\boldsymbol{\theta}_{11}^l) :\,& \big(\mathbbm{1}_{2} \otimes \mathbbm{1}_{2} \otimes \mathbbm{1}_{2} \big) \text{CCNOT} \big( \mathbbm{1}_{2} \otimes \mathbbm{1}_{2} \otimes \mathbbm{1}_{2} \big)  \\
    \hat{U}_{01}^l(\boldsymbol{\theta}_{01}^l) :\,& \big( \mathbbm{1}_{2} \otimes X\otimes \mathbbm{1}_{2}  \big) \text{CCNOT} \big(\mathbbm{1}_{2} \otimes X\otimes \mathbbm{1}_{2} \big)\\
    \hat{U}_{10}^l(\boldsymbol{\theta}_{10}^l) :\,& \big( \mathbbm{1}_{2} \otimes \mathbbm{1}_{2} \otimes X\big) \text{CCNOT} \big( \mathbbm{1}_{2} \otimes \mathbbm{1}_{2} \otimes X\big)\\
    \hat{U}_{00}^l(\boldsymbol{\theta}_{00}^l) :\,& \big( \mathbbm{1}_{2} \otimes X\otimes X\big) \text{CCNOT}\big( \mathbbm{1}_{2} \otimes X\otimes X\big)
    \label{mtl_gen}
\end{align}
where, $ \mathbbm{1}_{2} $ represents the identity gate on one qubit, and CCNOT denotes the Toffoli gate, which operates on $\ket{k}$ with control inputs $\ket{k_1}$ and $\ket{k_2}$ such that $k_1, k_2\in \{0,1\}$. We denote the single-qubit gates as $G(k_1, k_2)$, such that the general expression for the control of the transformation $\hat{U}_{k_1k_2}(\boldsymbol{\theta}_{k_1k_2})$ can be given by (from Eq.{~\ref{mtl_gen}}):
\begin{equation}
    \hat{U}_{k_1k_2}^l(\boldsymbol{\theta}_{k_1k_2}^l): G^l(k_1,k_2) \; \text{CCNOT} \; G^l(k_1,k_2) 
\end{equation}
where
\begin{align*}
    G^l(k_1,k_2) =& \big( \mathbbm{1}_{2} \otimes \left( k_1 \mathbbm{1}_{2} + (1-k_1) X \right) \\ 
    &\otimes \left( k_2 \mathbbm{1}_{2} + (1-k_2) X \right) \big).
\end{align*}

\subsubsection{K-assets case}
Furthermore, if we extend this logic to accommodate $K$ tasks, then we require $\log(K) + 1$ qubits, which is more than $\log(K)$, indicating the need for more qubits to represent $\ket{k}$. The resulting gate composition can then be expressed as:
\begin{align*}
   \hat{U}_{k_1k_2\hdots k_{logK}}^l(\boldsymbol{\theta}_{k_1k_2\hdots k_{logK}}^l) : &G^l(k_1,k_2\hdots k_{logK}) \; \text{CCNOT} \\
   & G^l(k_1,k_2\hdots k_{logK}) 
\end{align*}

where
\begin{align*}
    G^l(k_1,k_2\hdots k_{logK}) &= \big( \mathbbm{1} \otimes ( k_1 \mathbbm{1} + (1-k_1) X ) \\
    &\otimes ( k_2 \mathbbm{1} + (1-k_2) X ) \hdots \\
    & \otimes (k_{logK} \mathbbm{1} + (1 - k_{logK}) X )\big),
\end{align*} 
where $k_1,\hdots, k_{logK}\in\{0,1\}$. Note that this type of control provides exclusive task-specific layers for each task. In scenarios where a complete task-specific layer is not necessary for every task, the circuit can be optimized to reduce the number of gates by partially sharing the circuit among the tasks. Qubit overhead can also be reduced by using shared labels for similar assets.

\subsection{Training}

In classical MTL models, training typically involves optimizing weights using gradient-based methods applied to the entire network, with separate tasks handled through task-specific output layers or parameters. QMTL in this setting leverages shared parameters across tasks to capture common features while using task-specific parameters to model unique characteristics of each task. This approach often requires separate forward and backward passes for each task to compute gradients, making the process resource-intensive.

A parametric quantum circuit $ \hat{U}(\boldsymbol{\theta}) $ consists of fixed gates (such as CNOTs) and parameterized gates $\hat{G}^{lij}(\theta^{lij})=e^{-i\frac{\theta^{lij}}{2}\hat{P}^{lij}}$, where $\hat{P}^{lij}\in\{\hat{X}_i,\hat{Y}_i,\hat{Z}_i\}_{i=1}^{T+\tau}$ is a single qubit Pauli generator. The parametric quantum circuit consists of $m$ parameters, each corresponding to either a rotation gate $G_s^{lij}(\theta^{lij}_s)\otimes\mathbbm{1}_{K}$ (from Eq.~\eqref{s}) or a controlled rotation gate $\big( G_k^{lij}(\theta^{lij}_k) \otimes \ketbra{k}{k} + \mathbbm{1}_{2^{T+\tau}} \otimes \ketbra{k^\perp}{k^\perp} \big)$ (from Eq.~\eqref{ck}). However, during training, we fix the value of $k$, and since the label qubits are also excluded from measurement (see Figs.~\ref{stl_arc}, \ref{mtl_arc}, \ref{4mtl_arc}), we effectively reduce all the controlled rotation gates to $G_k^{lij}(\theta^{lij}_k)\otimes\mathbbm{1}_{K}$. This is because each task-specific layer is trained independently on its dataset, effectively making the control qubits function as a multiplexer. Mathematically, this leads to the equivalence of $\hat{U}^l_{ck}(\boldsymbol{\theta}^l_{ck})$ and $\hat{U}^l_{k}(\boldsymbol{\theta}^l_{k})$, resulting in
\begin{equation}
    \hat{U}(\boldsymbol{\theta}) = \prod_{l=1}^{L} \hat{U}_{k}^{l}(\boldsymbol{\theta}_k^{l}) \hat{U}_{s}^{l}(\boldsymbol{\theta}_{s}^l),
    \label{equivalence}
\end{equation}
where $\hat{U}(\boldsymbol{\theta})$ is now dimensionally reduced but retains the same structure as Eq.~\eqref{u} from the perspective of gradient computation. From Eq.\eqref{equivalence}, $\hat{U}(\boldsymbol{\theta})$ consists of a sequence of non-controlled unitary gates of size $2^{T+\tau}$. Consequently, gradient update rules from equations \eqref{shift} and \eqref{spsa} can be applied using the chain rule to train the QMTL model. This equivalence significantly reduces the complexity of the loss function and its dependence on hidden layers, resulting in training performance comparable to QSTL.

\subsection{Sequential Prediction}

\begin{figure}[t]
\centering
\includesvg[width=0.95\linewidth]{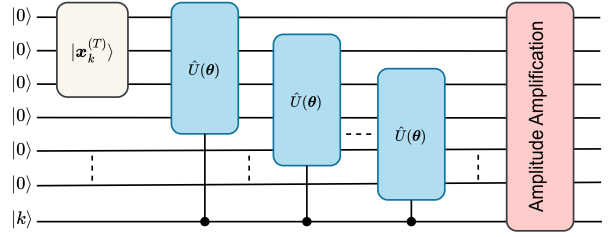}
\caption{\textbf{Sequential Prediction for an Asset.} A block diagram of our proposed sequential prediction approach, where a fully trained PQC ($\hat{U}(\boldsymbol{\theta})$) over $K$ stocks using quantum multi-task learning is used to predict $\tau=t$ ($\in\{1,\dots,R\}$) consecutive future values by loading the continuation on a new ancilla for each $t$. } 
\label{sequence}
\end{figure}

During inference, we load a sampled context from the overall contextual distribution but the circuit is capable of processing this single input to predict the future stock price for that specific context. For instance, we load a single context vector $ \boldsymbol{x}_k^{(T)} $ to a state $\ket{\boldsymbol{x}_k^{(T)}}$ and find (from Eq.~\eqref{y})
\[\ket{\boldsymbol{y}_k^{(T+1)}} \approx \sum_{{x^{T+1}_k \in \{ 0, 1 \}}}  \sqrt{\mathcal{P}(x_k^{T+1} | \boldsymbol{x}_k^{(T)}, \boldsymbol{\theta})} \ket{\boldsymbol{x}_k^{(T)}}\ket{x_k^{T+1}},\]
such that measuring the computational basis samples the most likely continuation based on the historical context given. However, we can repeatedly apply our QNN $R$ times, as shown in Fig.~\ref{sequence}, resulting in the final approximate state 
\begin{align*}
    \ket{\boldsymbol{y}_k^{(T+R)}} \approx \sum_{\boldsymbol{x}_{k}^{(T+R)/(T)} \in \{ 0, 1 \}^{R}} \hspace{-1em}&\Big(\sqrt{\mathcal{P}(\boldsymbol{x}_{k}^{(T+R)/(T)} | \boldsymbol{x}_k^{(T)}, \boldsymbol{\theta})} \nonumber\\
    &\ket{\boldsymbol{x}_k^{(T)}}\ket{x_k^{T+1}}\!\dots\!\ket{x_k^{T+R}}\Big).
    \label{repeat}
\end{align*}
This approach could, in future work, be extended to create a superposition state over all $d^R$ possible paths with logarithmic qubit depth $\mathcal{O}(R)$, allowing efficient encoding of highly nontrivial context-specific distributions over futures. Such an extension would enable the use of quantum algorithms for statistical tasks, potentially achieving quantum advantage at inference time, for example through quadratic sampling speedups using amplitude estimation for risk analysis~\cite{woerner2019quantum}. This is possible because QBGU preserves the mapping between input (context) and output (prediction) for individual samples within a batch, allowing the model to make predictions for each sample independently. This correspondence is further validated in the subsequent sections.

\begin{figure}[t]
\centering
\includesvg[width=\linewidth]{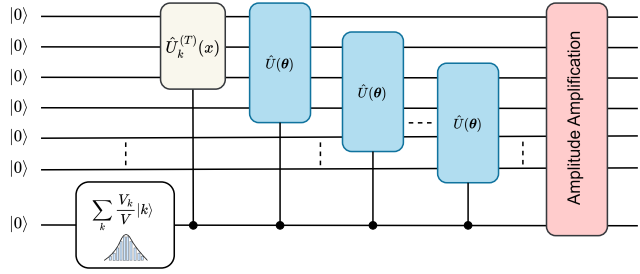}
\caption{\textbf{Sequential Prediction for a Portfolio of Assets.} A block diagram of our proposed sequential prediction approach, where a fully trained PQC ($\hat{U}(\boldsymbol{\theta})$) over $K$ stocks using quantum multi-task learning is used to predict $\tau=t$ ($\in\{1,\dots,R\}$) consecutive future values for an entire portfolio $ \sum_{k} \frac{V_k}{V} \ket{k} $.} 
\label{portsequence}
\end{figure}

In the QMTL setting, we can load portfolio distributions (such as market capitalization weighted index of publicly traded corporate stocks) to do sequence generation over the portfolio with only logarithmic overhead for the asset labels. Let $ V_{k} $ correspond to the weight of asset $ k $ and $ V = \sum_{k=1}^{K} V_{k} $. Then we can load the distribution $ \sum_{k=1}^{K} \sqrt{V_{k} / V} \ket{k} $ over the $ \log(K) $ label qubits and use label-control gates to initialize the corresponding context $ \ket{\boldsymbol{x}_{k}^{(T)}} $ dependent on each label. Applying our QNN $ R $ times prepares: 
\begin{align*}
\ket{\boldsymbol{y}^{(T+R)}} \approx \sum_{k=1}^{K} \sqrt{\frac{V_{k}}{V}} \ket{\boldsymbol{y}_{k}^{(T+R)}}. 
\end{align*}

Such a model can be utilized for quantum advantage on the downstream tasks, utilizing known quantum algorithms such as Ref.~\cite{woerner2019quantum}.

\section{Numerical Simulations}

\subsection{Loading Contextual Distribution}

To encode a contextual distribution, $\mathcal{P}(\boldsymbol{x}^{(T)})$ onto the wavefunction, we use a quantum circuit, illustrated in Fig.~\ref{loading}. For this experiment, we use historical stock data from Apple, discretized to binary values where $0$ indicates a price decrease, and $1$ indicates an increase. We choose a context length of $T=3$ and $\tau=1$, allowing the model to incorporate three consecutive days of stock data to predict one day into the future. Consequently, the contextual distribution is encoded using three qubits, each representing a day's information. Our dataset comprises $10,033$ stock prices, which we average weekly with a stride length of 1 day, yielding $10,029$ samples. 

\begin{figure}[t]
\centering
\includesvg[width=\linewidth]{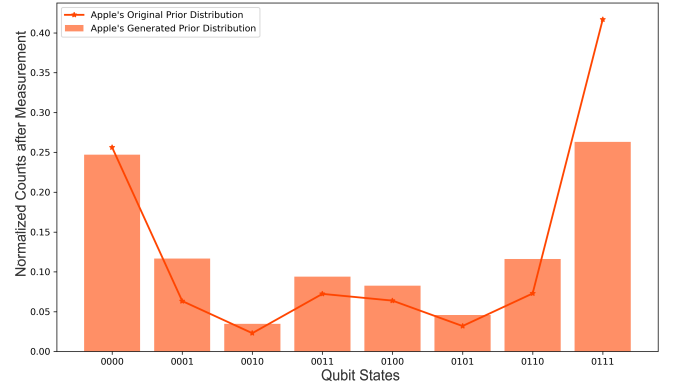}
\caption{\textbf{Contextual Probability Distribution.} A diagram showing the loaded contextual probability distribution ($\mathcal{P}(\boldsymbol{x}^{(T)})$) of Apple's stock data for a context size of $T=3$. } 
\label{prior}
\end{figure}

\begin{figure}[t]
\centering
\includesvg[width=\linewidth]{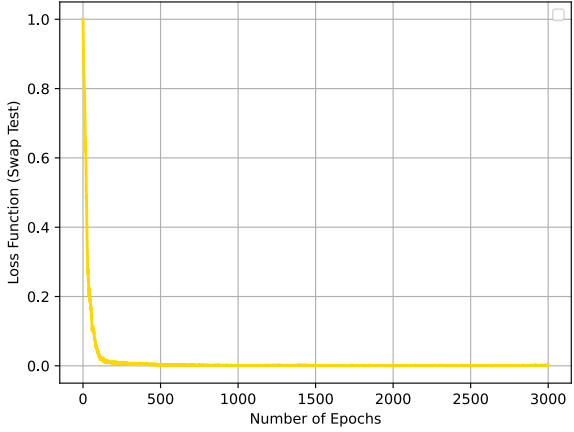}
\caption{\textbf{Training Loss for loading context distribution.} A diagram plotting the training loss for loading a contextual distribution ($\mathcal{P}(\boldsymbol{x}^{(T)})$) of Apple's stock data for a context size of $T=3$.} 
\label{prior_loss}
\end{figure}

This dataset is further divided into training ($80\%$) and testing datasets ($20\%$) by splitting. The quantum circuit includes a sequence of four layers ($L=4$) of parameterized quantum sub-circuits (also shown in Fig.~\ref{loading}). We employ the SPSA rule in conjunction with the swap test to iteratively adjust the weights, an approach whose advantages are discussed in later sections. The model was trained for 3000 epochs with a learning rate of 0.1. The number of measurement shots was set to 10,000, and the SPSA perturbation parameter was fixed at 0.01. All experiments were initialized using a random seed of 42 to ensure reproducibility. These parameters are the same for all the subsequent results in the manuscript unless otherwise stated. This results in a contextual distribution loaded as depicted in Fig.~\ref{prior}. The corresponding training loss is illustrated in Fig.~\ref{prior_loss}, demonstrating the model's convergence. As shown in Fig.~\ref{prior}, we observe that the contextual distribution is efficiently encoded onto the qubit states, accurately reflecting the intended probabilities.

\subsection{Quantum Single-Task Learning}

\begin{figure*}[t]
\centering
\includesvg[width=\linewidth]{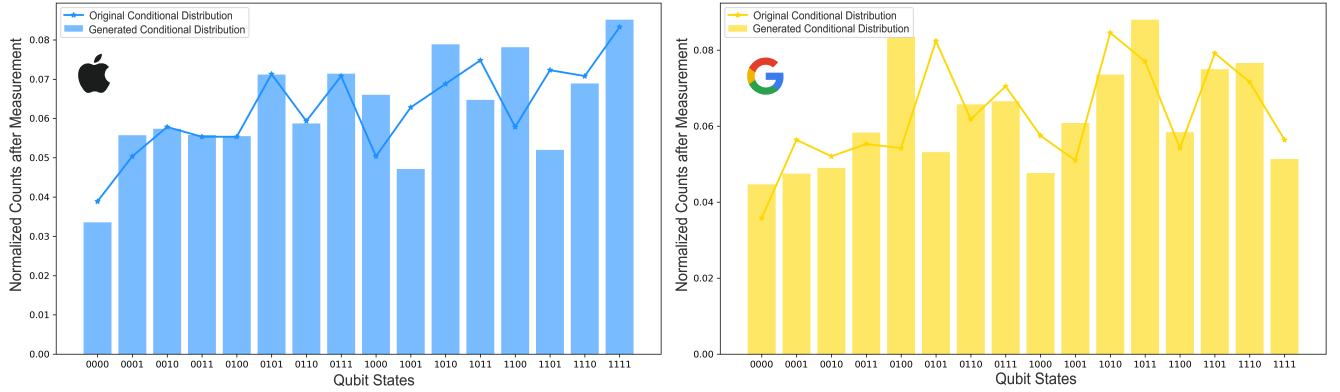}
\caption{\textbf{Predictions of Quantum Single-Task Learning for $T=3$.} Generated conditional probability distributions of both Apple and Google datasets are shown comparing them to the original distributions. The model was trained for $3000$ epochs with a learning rate of $0.1$.} 
\label{stl_plot}
\end{figure*}

With the contextual distribution now loaded, we proceed to predict the conditional probability $\mathcal{P}(x^{T+1}|\boldsymbol{x}^{(T)},\boldsymbol{\theta})$ from the contextual distribution $\mathcal{P}(\boldsymbol{x}^{(T)})$ using a quantum circuit as shown in Fig.~\ref{stl_arc}. To train this circuit, we use raw stock data from Apple and Google. The circuit employs eight qubits: the first three qubits store the contextual distribution loaded in the previous section, and a fourth qubit is designated for the prediction. The remaining four qubits encode the target conditional probability distribution $\mathcal{P}(x^{T+1}|\boldsymbol{x}^{(T)},\boldsymbol{\theta})$, loaded similarly to the contextual. The circuit incorporates four layers ($L=4$) of parameterized gates, which upon training, generate the wavefunction $\ket{\boldsymbol{y}^{(T+1)}}$, representing the predicted return for the context input $\ket{\boldsymbol{x}^{(T)}}$. To optimize this prediction, we apply the swap test, measuring the distance between the original and predicted conditional probabilities. This distance metric guides the training process by yielding gradients via the SPSA rule, allowing for precise adjustment of $\boldsymbol{\theta}$ to minimize prediction error and improve model accuracy. We trained the model for $3000$ epochs with a learning rate of $0.1$ to get the predicted conditional probability (or distributions) as displayed in Fig.~\ref{stl_plot}.

The distance between the target conditional probability distribution and the predicted conditional probability distribution is quantified using the KL Divergence, providing a measure of similarity between the two distributions. These values are detailed in Table 1, offering insight into the model's accuracy and convergence during training. Lower KL Divergence values indicate closer alignment with the target distribution, demonstrating the effectiveness of our quantum circuit in capturing the underlying patterns of stock price movements.

\subsection{Quantum Multi-Task Learning}

\begin{figure*}[h]
\centering
\includesvg[width=\linewidth]{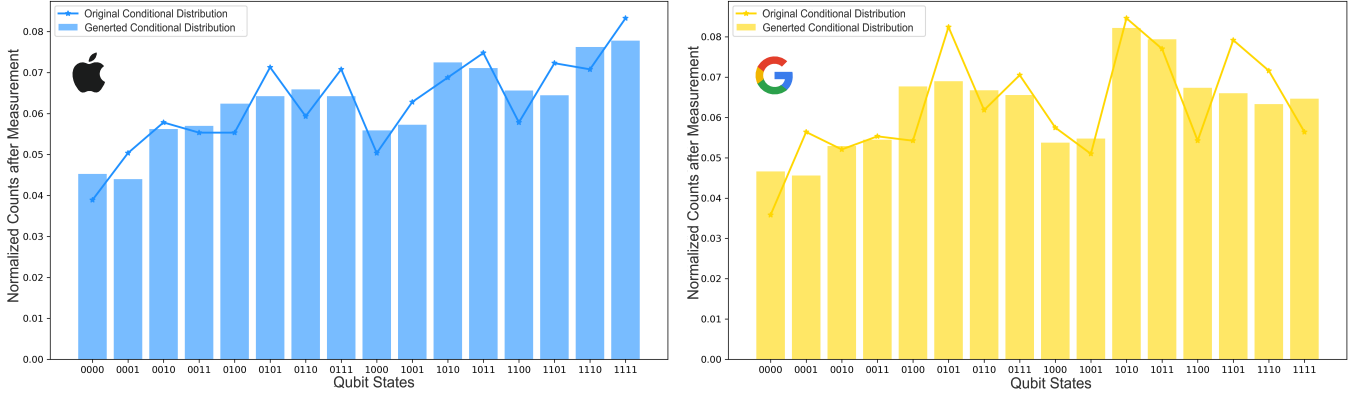}
\caption{\textbf{Predictions of Quantum Multi-Task Learning for $T=3$ and $K=2$}. Generated conditional probability distributions of both Apple and Google datasets are shown comparing them to the original distributions. The model was trained for $3000$ epochs with a learning rate of $0.1$.} 
\label{mtl_plot}
\end{figure*}

To train multiple stocks simultaneously, we employ a multi-task learning architecture, illustrated in Fig.~\ref{mtl_arc}, using both Apple and Google stock data. The circuit uses a control qubit, $\ket{k}$, to distinguish between the stocks, where
$\ket{k}=\ket{0}$ represents Apple, and $\ket{k}=\ket{1}$ represents Google. We utilize a single set of parameterized gates, including $\hat{U}_s(\boldsymbol{\theta}_s)$, $\hat{U}_1^1(\boldsymbol{\theta}_1^1)$, and $\hat{U}_2^1(\boldsymbol{\theta}_2^1)$, running the circuit for 250 epochs per stock. Training begins with Apple ($\ket{k}=\ket{0}$) for $250$ epochs, followed by Google ($\ket{k}=\ket{1}$) for another 250 epochs, both at a learning rate $\beta= 0.1$. The predicted conditional probability functions are shown in Fig.~\ref{mtl_plot}, where we observe a closer alignment to the target distributions compared to single-task learning. Table 1 further confirms this, showing that the KL Divergences in the multi-task learning case are lower. We attribute this improvement to correlations captured between the Apple and Google datasets, which induce a regularization effect on the circuit. This shared representation, attributed to the shared parameters $\hat{U}(\boldsymbol{\theta})$, allows the model to retain information from both stocks, creating an inductive bias that enhances performance.

It is important to note, however, that classical deep learning models can outperform quantum approaches in conventional large-scale time-series prediction tasks. This superiority arises from the scalability of classical architectures, which can accommodate millions of parameters, deep attention stacks, and long temporal contexts beyond the reach of near-term quantum systems. In such scenarios, Transformers and LSTMs achieve state-of-the-art performance in purely classical forecasting pipelines. Nevertheless, in this work, time-series prediction accuracy is not the sole key performance indicator. Our emphasis lies in efficient realization on quantum hardware, achieved through the QBGU mechanism, and enable sequential prediction as discussed in Section V. The proposed QMTL with QBGU framework can be repeatedly applied to predicted future stock states in superposition, allowing the exploration of all alternate future trajectories simultaneously. Such a process, where probabilistic futures are evolved and measured in parallel, remains infeasible for classical models, which must evaluate each scenario sequentially.

\begin{table}[h!]
\centering
\caption{\textbf{Performance of QSTL vs QMTL for two selected stocks.} }
\begin{tabular}{|c|c|c|c|} 
\hline
 Asset & Model & Parameters & KL Divergence  \\
 \hline
 Apple & STL & 32 & 0.1047 \\
 \hline
 Google & STL & 32 & 0.1239 \\
 \hline
 Apple & MTL & 16 & 0.0614 \\
 \hline
 Google & MTL & 16 & 0.0754 \\
 \hline
\end{tabular}\\
\label{tab}
\end{table}

\begin{figure*}[h]
\centering
\includesvg[width=\linewidth]{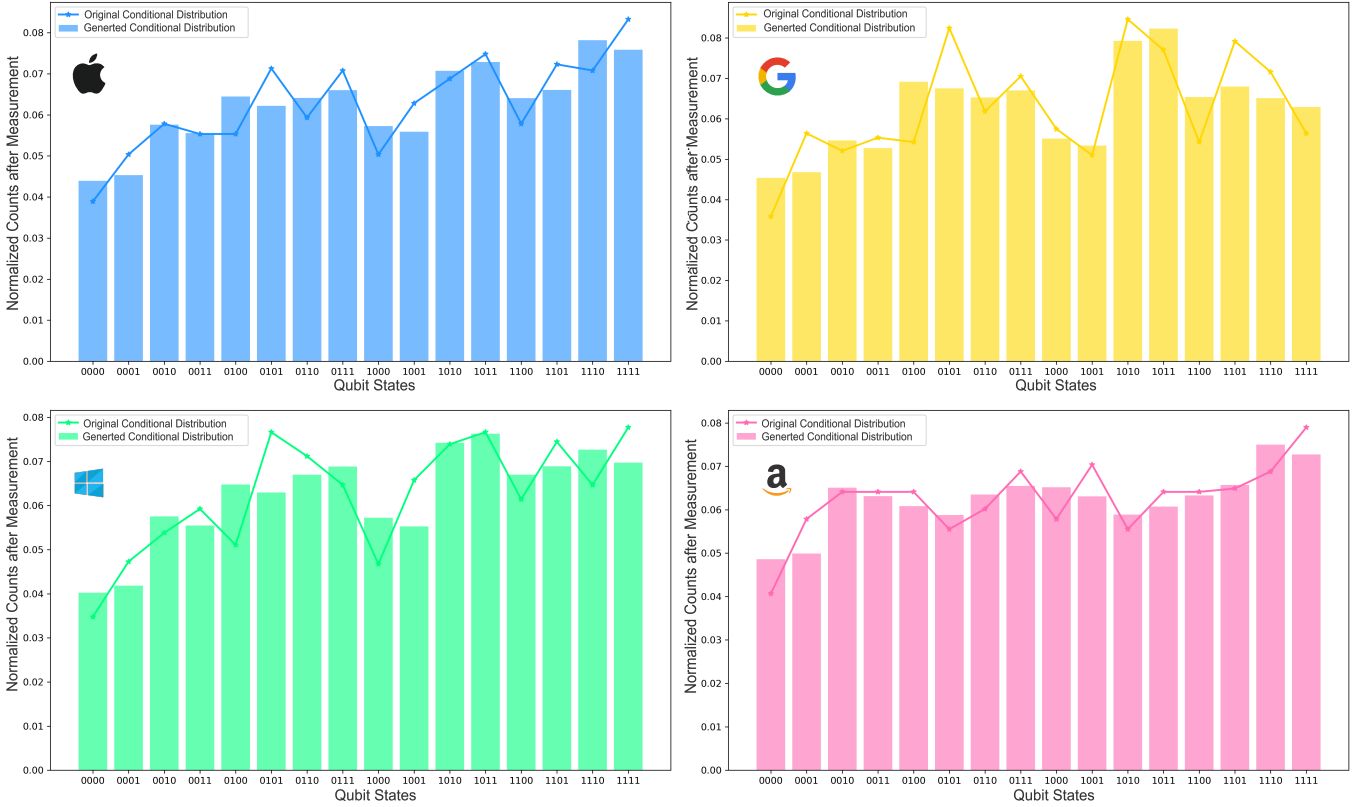}
\caption{\textbf{Predictions of Quantum Multi-Task Learning for $T=3$ and $K=4$}. Generated conditional probability distributions of Apple, Google, Microsoft, and Amazon datasets are shown comparing them to the original distributions. The model was trained for $3000$ epochs with a learning rate of $0.1$.} 
\label{4mtl_plot}
\end{figure*}

\subsubsection{Convergence}

\begin{figure}[t]
\centering
\includesvg[width=\linewidth]{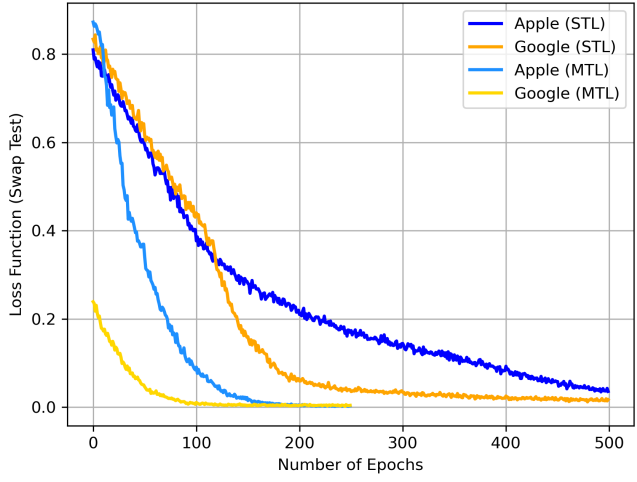}
\caption{\textbf{Loss over training epochs for QSTL and QMTL}. A diagram plotting loss function versus epochs for QMTL and QSTL.} 
\label{mtl_loss}
\end{figure}

The training losses for both models, QMTL and QSTL, are plotted in Fig.~\ref{mtl_loss}. We observe that the QMTL model converges significantly faster than the QSTL model, reflecting its improved learnability due to correlations across datasets. Notably, the QMTL loss curve for Google begins at a lower starting point. This is due to the sequential training, where Apple’s dataset is processed first, followed by Google’s. By the time Google’s data is introduced, the model has already incorporated information from Apple, and the correlation between the stock prices of Apple and Google leads to a reduced initial loss for Google. This demonstrates how QMTL effectively leverages inter-dataset correlations to enhance learning efficiency and stability.

%
%
%
%
%
%
%
%
%
%
%
%
%
%
%
%
%
%
%
%
%
%
%
%
%
%
%
%
%
%
%
%
%

\subsubsection{SWAP Test versus Mean-Squared Error Loss}

\begin{figure}[t]
\centering
\includesvg[width=\linewidth]{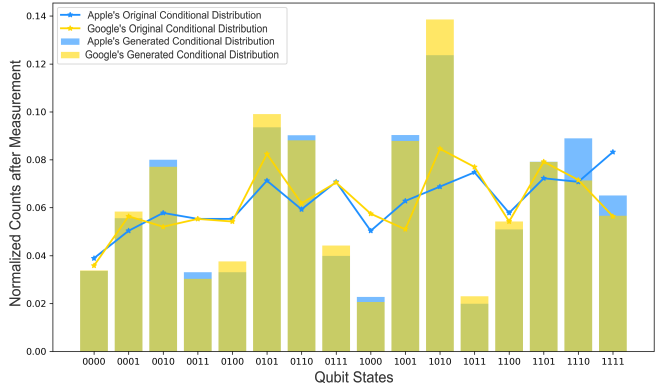}
\caption{\textbf{MSE Loss.} Representing the underlying conditional probability distributions of Apple and Google stocks after training with MSE Loss.} 
\label{mse}
\end{figure}

The motivation for using SPSA in optimizing a quantum neural network is well-established as it provides a gradient estimation technique compatible with quantum circuits, allowing efficient optimization by leveraging the circuit's differentiable structure without requiring classical backpropagation. Furthermore, we examine the differences between using the swap test and MSE loss in conjunction with SPSA. To investigate, we train two QSTL models on Apple and Google datasets, using MSE loss for $10,000$ epochs with a learning rate of $0.00001$. The resulting predicted conditional probabilities are plotted in Fig.~\ref{mse}, where we observe that MSE loss paired with SPSA fails to capture the target distribution accurately. This limitation likely arises from the non-trigonometric nature of MSE, which does not align well with the periodicity inherent in quantum operations. Consequently, we exclusively utilized the swap test as the loss function throughout this paper to ensure optimal distribution learning. From the Fig.~\ref{mse} and Fig.~\ref{stl_plot}, we can conclude that our novel training method to exploit a conditionalized fidelity loss over quantum distributions shows significantly better performance compared to standard measurements for computing the MSE loss.

%
%
%

%
%

%
%


%
%

\subsubsection{Scalability}
To explore the scalability of QMTL, we expand the model to learn four stocks: Apple, Google, Microsoft, and Amazon. Using the circuit in Fig.~\ref{4mtl_arc}, we get $3$ control qubits, of which only two qubits ($\ket{k_1}$ and $\ket{k_2}$) are externally controlled. The circuit consists of one repetition of the trainable shared circuit $\hat{U}_s(\boldsymbol{\theta}_s)$, followed by the task-specific circuits $\hat{U}_{11}^1(\boldsymbol{\theta}_{11}^1)$, $\hat{U}_{01}^1(\boldsymbol{\theta}_{01}^1)$, $\hat{U}_{10}^1(\boldsymbol{\theta}_{10}^1)$, and $\hat{U}_{00}^1(\boldsymbol{\theta}_{00}^1)$. We train the model similarly to the two-stock case, with 250 epochs per stock and the same learning rate. The resulting predicted conditional probability distributions, plotted in Fig.~\ref{4mtl_plot}, demonstrate that the model effectively learns the distributions for all four stocks while using the same number of trainable parameters as in the two-stock scenario. This scalability illustrates the robustness of the QMTL approach, which benefits from efficient parameter sharing and enhanced correlation capture, thereby outperforming single-task learning even as the number of assets grows.

\begin{figure}[t]
\centering
\includesvg[width=\linewidth]{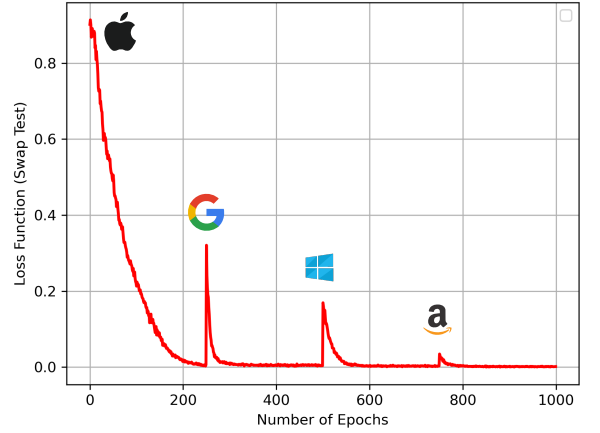}
\caption{\textbf{QMTL Loss over training epochs of 4 assets.} A diagram plotting loss function vs. epochs for QMTL ($K=4$) with a learning rate of $0.1$.} 
\label{4mtl_loss}
\end{figure}

Fig.~\ref{4mtl_loss} shows the loss function progression during the training of all stocks, illustrating the model's adaptability to each dataset. Each peak aligns with transitions between datasets, and we observe a gradual reduction in peak heights over time. This decline suggests that the model is capturing correlations between stocks, enhancing its compatibility across different datasets and resulting in consistently lower loss values. This trend underscores the QMTL model's ability to leverage shared patterns, improving overall training stability and efficiency across multiple assets. 

\begin{figure}[t]
\centering
\includesvg[width=\linewidth]{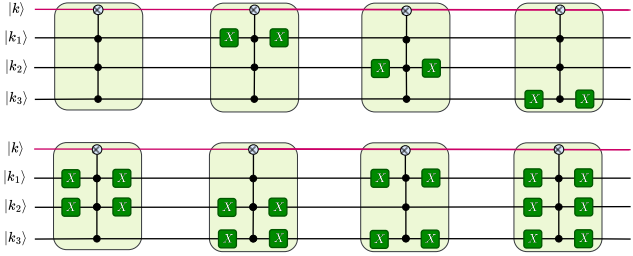}
\caption{\textbf{QMTL Control Circuit for K=8 assets.} Share-and-specify ansatz is not shown for simplicity.}
\label{8mtl_circuit}
\end{figure}

To further evaluate the scalability and generalization capacity of the QMTL framework, we extend the architecture to simultaneously learn eight assets: Apple, Google, Microsoft, Amazon, Pepsi, Western Digital, Texas Instruments, and IBM. We average the asset prices over 3 days instead of 5 (weekly) for variability. This configuration expands the control space to four control qubits, of which three qubits ($\ket{k_1}$, $\ket{k_2}$, and $\ket{k_3}$) are externally modulated to select among eight task-specific circuits as shown in Fig.~\ref{8mtl_circuit}. The circuit structure remains analogous to the previous $K=4$ setup.

\begin{figure*}[!t]
\centering
\includesvg[width=\linewidth]{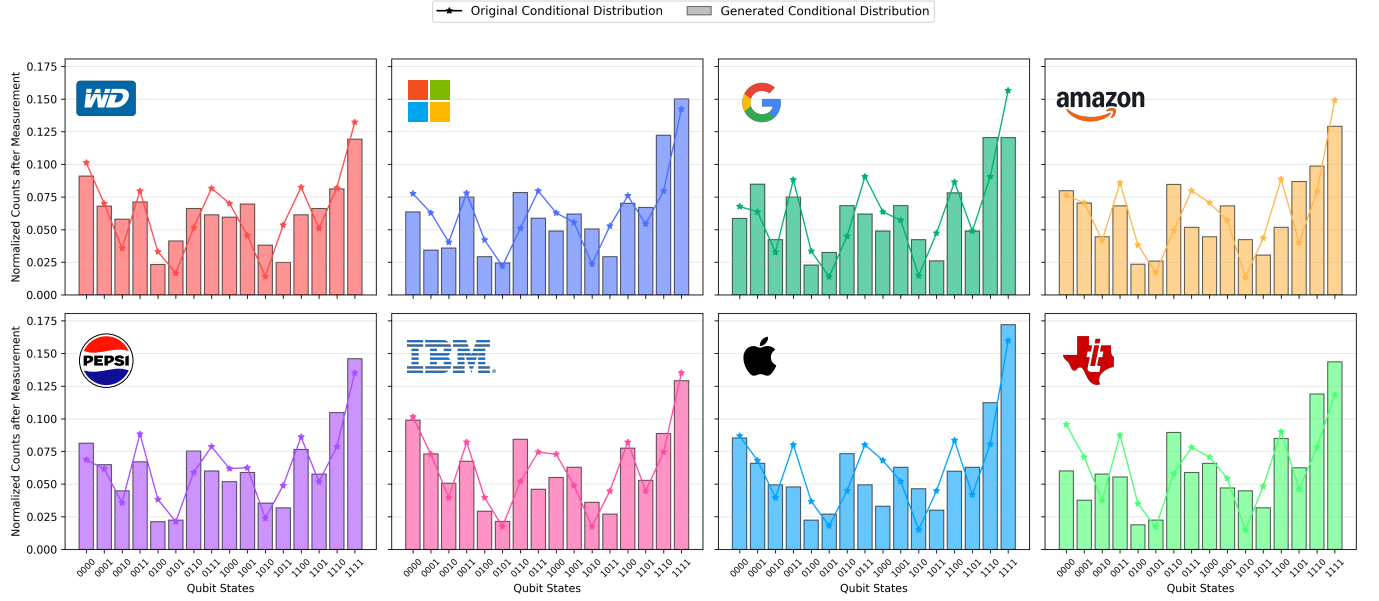}
\caption{\textbf{Predictions of Quantum Multi-Task Learning for $T=3$ and $K=8$}. Generated conditional probability distributions of various datasets are shown comparing them to the original distributions. The model was trained for $800$ epochs with a learning rate of $0.1$.} 
\label{8mtl}
\end{figure*}

We train the model for 100 epochs per stock and the same learning rate. The predicted conditional probability distributions, shown in Fig.~\ref{8mtl}, confirm that the model accurately learns all eight stocks while keeping the number of trainable parameters the same. Each distribution closely matches its target, showing that QMTL maintains good prediction quality even as the number of tasks increases. The shared circuit continues to capture common patterns across different stocks, while each task-specific part learns the unique behavior of an individual stock. Overall, the eight-stock results highlight the robustness and efficiency of the QMTL model in handling larger and more complex multi-asset learning problems.

\begin{figure}[t]
\centering
\includegraphics[width=\linewidth]{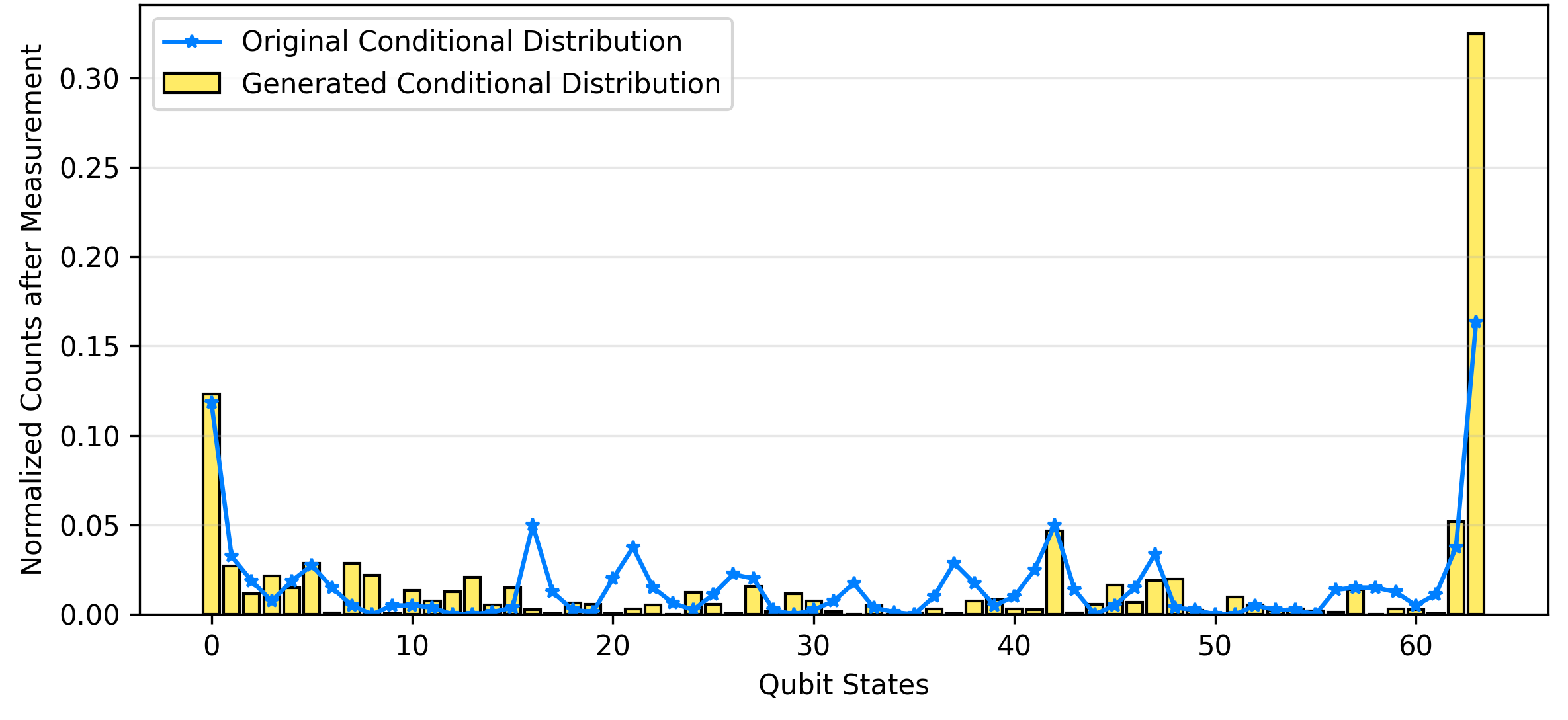}
\caption{\textbf{Predictions of QMTL for $T=2$ and $d=4$.} Generated conditional probability distribution of Apple dataset is shown comparing it to the original distribution. The model was trained for $3000$ epochs with a learning rate of $0.1$.}
\label{quant4_plot}
\end{figure}

To further explore how QMTL behaves under richer data representations, we conducted an additional experiment where stock prices were quantized into four discrete levels instead of two. This setting provides finer granularity in capturing subtle variations in stock movements while maintaining a manageable circuit size. Rather than dividing the range of returns into evenly spaced bins, we employed a non-uniform, density-based quantization scheme in which the boundaries between quantization levels are determined by the empirical distribution of the data. In this approach, the full range of price changes is partitioned according to their cumulative probability density so that each quantization level contains approximately the same proportion of data points. This means that smaller intervals are assigned to regions where the data is dense, while larger intervals cover the sparse, tail regions where extreme price changes are rare. Such a representation ensures that all quantization levels are statistically meaningful and that the model allocates more expressive capacity to the most frequently occurring variations.

For this experiment, we focused on a context size of two ($T=2$), rather than three, to keep the results intuitive and the number of possible qubit states within a reasonable range. The circuit is similar to Fig.~\ref{mtl_arc} but context and prediction bits now correspond to two qubits each. As shown in Fig.~\ref{quant4_plot}, the predicted and original conditional probability distributions align closely, indicating that the model effectively captures the underlying statistical structure of the quantized data.

\begin{figure}[t]
\centering
\includegraphics[width=\linewidth]{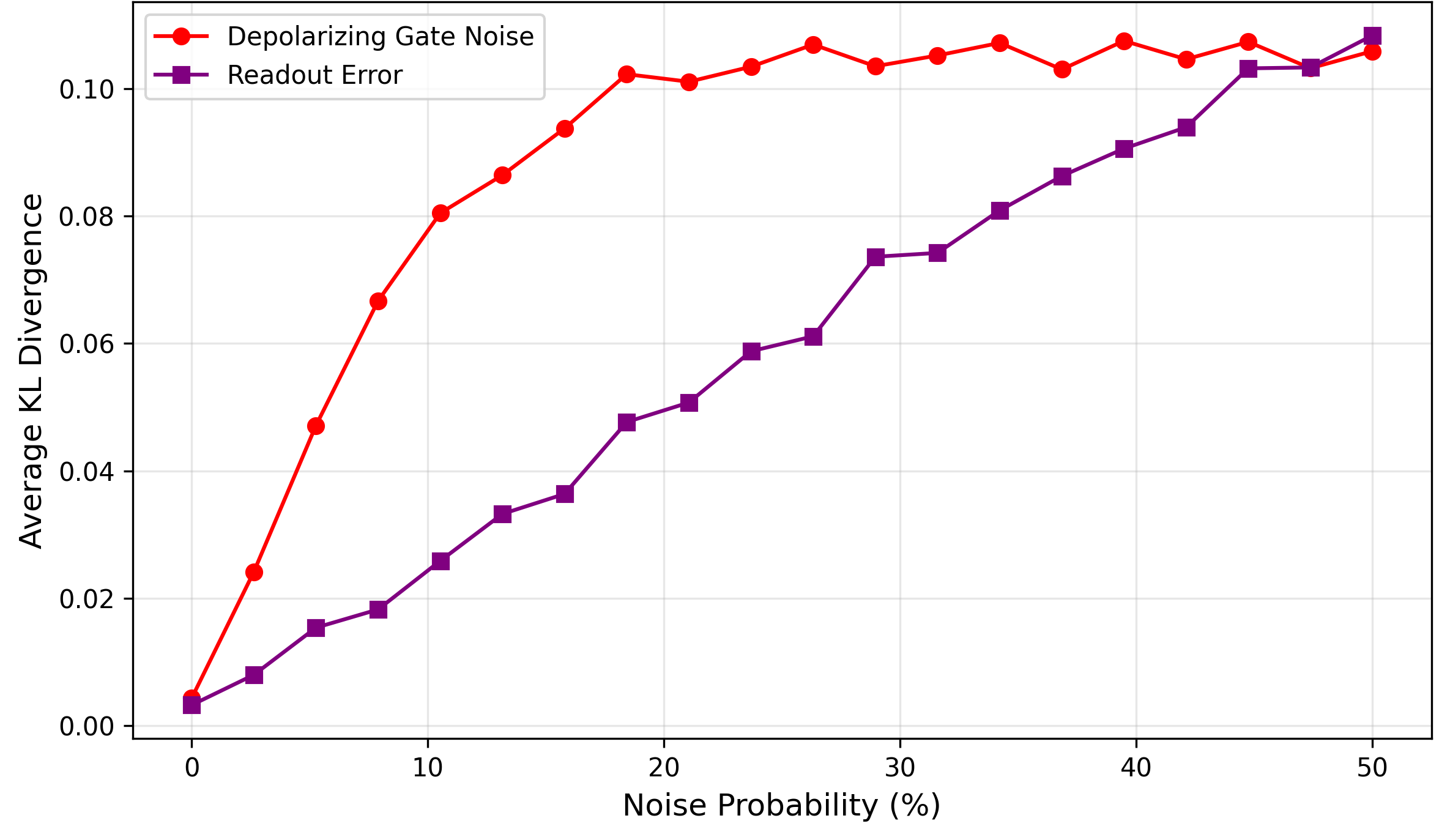}
\caption{\textbf{Effect of Depolarizing Noise and Readout Error on QMTL.} Measured KL divergences between noisy and noiseless output distributions from the QTML model for Apple stock prices with varying noise probabilities.}
\label{noise}
\end{figure}

\subsubsection{Noise}
To study how hardware imperfections affect the model, we used Qiskit’s AerSimulator to test QMTL on a single stock under different levels of depolarizing gate noise and readout error. Figure~\ref{noise} shows the resulting average KL divergence between the noisy and ideal output distributions as the noise probability increases from 0 to 50\%.

Both noise models were implemented using Qiskit’s built-in noise framework. For depolarizing noise, the noise probability represents how likely it is that a qubit’s state is randomly disturbed during a gate operation, replacing it with a completely mixed state \cite{depolarizing_noise}. For readout error, the noise probability indicates how often a measured bit is flipped during measurement \cite{readout_error}. Note that there are other types of noise prevalent in quantum circuits which are extensively studied in literature \cite{noise}.

As the figure shows, both noise sources cause the KL divergence to increase as the noise becomes stronger, meaning that the output distribution moves further away from the noiseless reference. Depolarizing noise (red curve) shows a faster and larger rise in divergence, reaching around 0.1 at high noise levels, while readout error (purple curve) increases more gradually and almost linearly. This difference reflects the fact that gate noise affects the computation throughout the circuit, whereas readout noise influences only the final measurement.

\section{Conclusion}

In this paper, we presented a quantum multi-task learning (QMTL) architecture tailored for predicting stock prices distribution across multiple assets. By encoding contextual distributions into quantum states and leveraging a compact, parameter-efficient circuit, our approach capitalizes on shared patterns across stocks, resulting in enhanced accuracy and faster convergence compared to quantum single-task learning (QSTL). We optimized our training using quantum batch gradient update (QBGU) that enabled training over a distribution of inputs. The scalability of the QMTL model was validated by extending predictions to multiple stocks with the same circuit, showing its potential for broader financial applications. Our results underscore the viability of quantum machine learning for loading complex financial distributions, paving the way for future studies in quantum finance that harness the unique capabilities of multi-task quantum neural networks for more efficient and adaptive forecasting models and the utilization of such networks for downstream quantum prediction that enable quantum advantage.

\section{Acknowledgement}

The authors thank Sarvagya Upadhyay, Yasuhiro Endo, and Hirotaka Oshima for their thoughtful comments. The authors thank anonymous reviewers for their insightful and constructive feedback.

\section{Data Availability}

Data generated in this study are available from the corresponding author upon reasonable request.

\appendix*

\end{document}